\documentclass[11pt]{article}

\usepackage[preprint]{acl}

\usepackage{times}
\usepackage{latexsym}
\usepackage{graphicx}
\usepackage{amsmath}
\usepackage{amsthm}
\usepackage{amsfonts}
\usepackage{booktabs}
\usepackage{algorithm}
\usepackage{algorithmic}
\usepackage[switch]{lineno}
\usepackage{natbib}
\usepackage{makecell}
\usepackage{multirow}
\usepackage{subcaption}
\usepackage{pifont}
\usepackage{xcolor}     
\usepackage[table]{xcolor}
\usepackage{amssymb} 
\usepackage{listings}
\usepackage{enumitem}
\usepackage{pgf}
\usepackage{array}
\usepackage{xfp}
\usepackage{verbatim}
\usepackage{stfloats}
\usepackage{placeins}
\usepackage{titletoc}      

\hypersetup{
  colorlinks=true,
  linkcolor=blue!50!black,   
  linktoc=all
}

\lstdefinestyle{promptstyle}{
  basicstyle=\small\ttfamily,
  breaklines=true,
  breakatwhitespace=false,
  columns=fullflexible,
  keepspaces=true,
  frame=none,
  xleftmargin=0pt,
  aboveskip=2pt,
  belowskip=2pt,
}

\definecolor{scoreBlue}{HTML}{4575A8}
\definecolor{scoreRed}{HTML}{B05252}

\newcommand{\ScoreLower}{0.5}

\newcommand{\scorecell}[1]{%
  \edef\scoreVal{\fpeval{min(max(#1,0),1)}}%
  \edef\scoreMid{\fpeval{(\ScoreLower+1)/2}}%
  \ifdim\scoreVal pt<\scoreMid pt
    \edef\scoreMix{\fpeval{round(100*(\scoreMid-\scoreVal)/(\scoreMid-\ScoreLower),0)}}%
    \edef\scoreMix{\fpeval{min(max(\scoreMix,0),100)}}%
    \edef\scoreBg{scoreRed!\scoreMix!white}%
    \ifdim\scoreMix pt>60pt
      \expandafter\cellcolor\expandafter{\scoreBg}\textcolor{white}{#1}%
    \else
      \expandafter\cellcolor\expandafter{\scoreBg}\textcolor{black}{#1}%
    \fi
  \else
    \edef\scoreMix{\fpeval{round(100*(\scoreVal-\scoreMid)/(1-\scoreMid),0)}}%
    \edef\scoreMix{\fpeval{min(max(\scoreMix,0),100)}}%
    \edef\scoreBg{scoreBlue!\scoreMix!white}%
    \ifdim\scoreMix pt>60pt
      \expandafter\cellcolor\expandafter{\scoreBg}\textcolor{white}{#1}%
    \else
      \expandafter\cellcolor\expandafter{\scoreBg}\textcolor{black}{#1}%
    \fi
  \fi
}

\newtheorem{definition}{Definition}
\newtheorem{axiom}{Axiom}

\newtheorem{proposition}{Proposition}

\newcommand*{\QEDB}{\null\nobreak\hfill\ensuremath{\square}}%

\usepackage{csquotes}
\MakeOuterQuote{"}  

\usepackage[export]{adjustbox}

\usepackage[T1]{fontenc}

\usepackage[utf8]{inputenc}

\usepackage{microtype}

\usepackage{inconsolata}

\usepackage{graphicx}

\usepackage{xcolor}
\definecolor{graybg}{gray}{0.92}
\definecolor{ForestGreen}{RGB}{34, 139, 34}
\definecolor{colorA}{HTML}{C0392B} 
\definecolor{colorB}{HTML}{2980B9}
\definecolor{crimson}{HTML}{DC143C}

\newcommand\cww[1]{{\textcolor{black}{#1}}}
\newcommand\tj[1]{{\textcolor{black}{#1}}}
\newcommand\cw[1]{{\textcolor{black}{#1}}}
\newcommand\yf[1]{{\textcolor{black}{#1}}}

\newcommand\cam[1]{{\textcolor{black}{#1}}}

\usepackage[most]{tcolorbox}
\usepackage{listings}

\newtcblisting{promptbox}[1]{
    enhanced,
    breakable,
    listing only,
    colback=white,
    colframe=blue!70!black,
    colbacktitle=blue!85!black,
    coltitle=white,
    fonttitle=\bfseries\normalsize,
    title={#1},
    boxrule=0.6pt,
    arc=0pt,
    outer arc=0pt,
    left=1.5mm,
    right=1.5mm,
    top=1mm,
    bottom=1mm,
    listing options={
        basicstyle=\ttfamily\small,
        breaklines=true,
        columns=fullflexible,
        keepspaces=true,
        showstringspaces=false
    }
}

\title{\textit{Grammar of the Wave}: Towards Explainable Multivariate Time Series\\ Event Detection via Neuro-Symbolic VLM Agents}

\author{
  \textbf{Sky Chenwei Wan\textsuperscript{1,2}}, \textbf{Yifei Y. Wang\textsuperscript{2}}\thanks{Work done during Yifei's internship at SLB.}, \textbf{Tianjun Hou\textsuperscript{1}}, \textbf{Xiqing Chang\textsuperscript{1}}, \textbf{Aymeric Jan\textsuperscript{1}}\\
  \textsuperscript{1}AI Lab, SLB, France\\
  \textsuperscript{2}LTCI, T\'el\'ecom Paris, Institut Polytechnique de Paris, France\\
  \texttt{\href{cwan5@slb.com}{cwan5@slb.com}}, \texttt{\href{thou2@slb.com}{thou2@slb.com}}
}

\begin{document}
\maketitle
\begin{abstract}
Time Series Event Detection (TSED) aims to localize semantically meaningful events in time series data, with critical applications in high-stakes domains. Unlike statistical anomalies, events are often defined by natural-language descriptions with internal temporal-logic structures across multiple physical channels. However, in real-world settings, dense event annotations are expensive to obtain, making purely supervised learning difficult. We introduce \textbf{Language-guided TSED}, a setting where a model is given textual event descriptions and must ground them to intervals in multivariate signals with little or no labeled data. To address this problem, we propose \textbf{Event Logic Tree} (ELT), a knowledge representation framework that converts linguistic descriptions into structured temporal logic over signal primitives. Building on ELT, we present \textsc{SELA}, a neuro-symbolic VLM agent framework that iteratively grounds primitives from signal visualizations and composes them under ELT constraints, producing both event intervals and faithful tree-structured explanations. We further release a real-world benchmark across energy and climate domains with expert knowledge and annotations. Experiments show that \textsc{SELA} improves over supervised fine-tuning and existing zero/few-shot time series reasoning baselines\footnote{Our project is available at \url{https://github.com/cw-wan/SELA}.}.
\end{abstract}

\section{Introduction}
\label{sec:intro}

Time Series Event Detection aims to localize temporal segments in multivariate signals where specific events occur. It is important in high-stakes domains such as health~\citep{akara-2017-deepsleepnet, perslev-2019-utime} and energy production~\citep{khaouja-2025-do}. Unlike time series classification, which assigns a global label to a sequence~\citep{classification-survey}, event detection requires fine-grained localization at the segment level. It is also different from anomaly detection, which mainly targets statistical deviations~\citep{anomaly-detection-survey}. Events are semantically defined processes, often involving cross-channel temporal relations, such as \textit{``a sharp rise on channel A followed by a dip on channel B''}.

\begin{figure}[ht]
\centering
\scalebox{0.98}{
\includegraphics[width=\columnwidth]{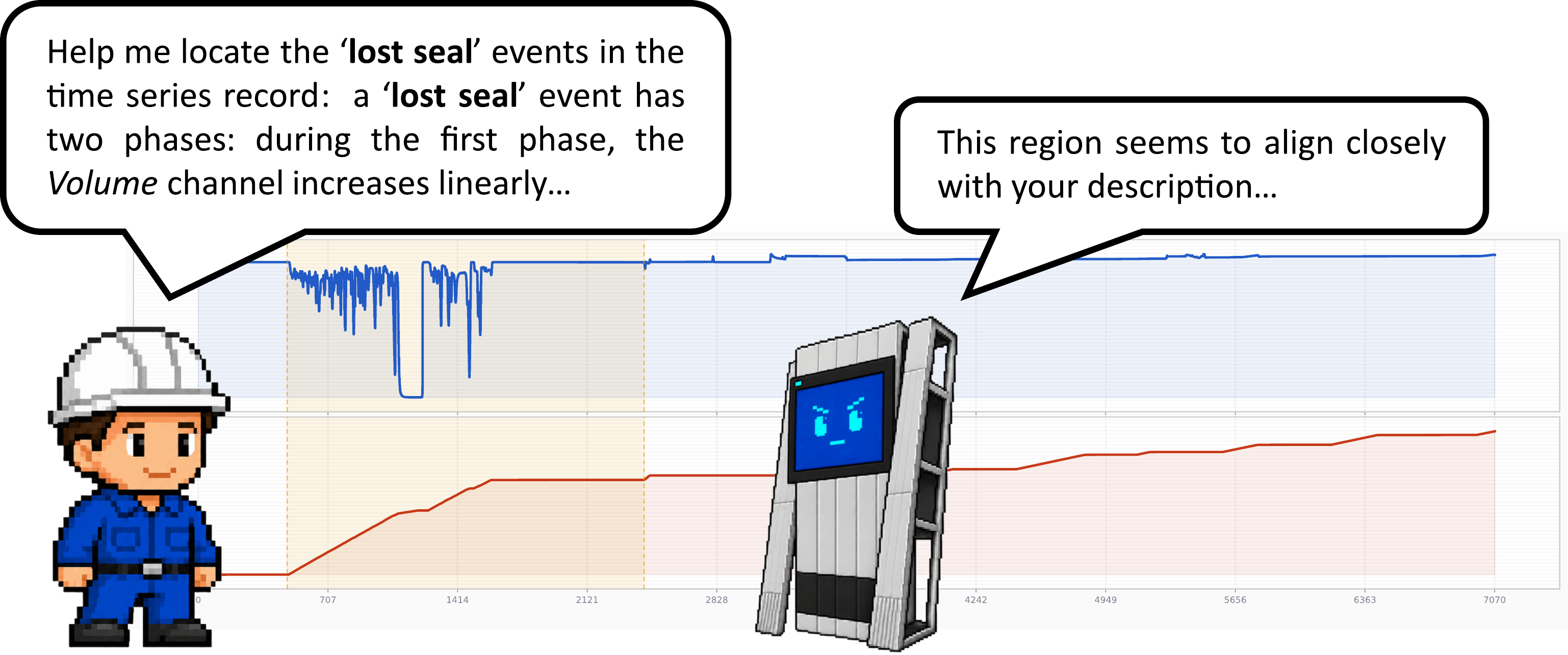}
}
\caption{An illustration of the Language-guided Time Series Event Detection task.}
\label{fig:task}
\end{figure}

Most existing approaches rely on supervised learning, including CNN or Transformer-based models (\citealp{perslev-2019-utime,timesnet}) and fine-tuning time series foundation models \citep{khaouja-2025-do}. However, in many industrial and scientific settings, dense event annotations are expensive to obtain, while textual event descriptions are often available from operational manuals or expert guidelines. This motivates a \textbf{language-guided setting}: given a multivariate time series and textual descriptions of target events, the model should ground the descriptions into event intervals without large amounts of labeled data (Figure \ref{fig:task}).

The central challenge is \textit{grounding}. Textual descriptions are unstructured, such as the following description from the energy domain:

\begin{quote}
During the \textbf{Buildup} event, the \textit{Volume} should remain stable across the entire phase. \underline{Simultaneously}, the \textit{Pressure} should \underline{either} recover slightly with the rate of change decreasing \underline{or} bounce up quickly, \underline{and then} reach a steady phase.
\end{quote}

This description is not a flat label. It defines atomic signal patterns, such as \textit{Volume remains stable} and \textit{Pressure recovers slightly}, and relates them through temporal-logic cues such as \textit{simultaneously}, \textit{either}, and \textit{and then}. Therefore, an explainable detector should not only predict the global event interval, but also ground its sub-patterns to concrete signal segments as verifiable evidence.

To address this challenge, we introduce \textbf{Event Logic Tree} (ELT), a knowledge representation framework that "translate" textual event descriptions into tree-structured temporal logic. ELT decomposes an event into atomic signal primitives and composes them through operators such as succession, synchronization, containment, and alternatives. Based on ELT, we build \textsc{SELA}, a neuro-symbolic agent system for zero-shot language-guided event detection. A \textit{Logic Analyst} parses the textual description into an ELT schema, while a \textit{Signal Inspector} grounds the schema in the observed time series by locating and refining event intervals. ELT serves as the central \textit{intermediate representation}: it guides the grounding process and produces an instantiated proof tree that can be inspected by human experts.

To evaluate our approach, we release \textsc{KITE}, a real-world benchmark with expert-verified event descriptions and event annotations from energy and climate domains. Experiments show that \textsc{SELA} consistently improves event grounding and precise localization, demonstrating the benefit of ELT-guided reasoning.

\paragraph{}Our contributions are summarized as follows:

\begin{itemize}
    \item We formulate a language-guided setting for detecting semantically defined events in multivariate time series data.
    \item We propose \textbf{ELT}, a structured representation that helps faithfully ground textual event descriptions into time series intervals with interpretable primitive-level evidence. Based on this, we develop \textbf{\textsc{SELA}}, a neuro-symbolic agent system that uses ELT to perform zero-shot event semantic grounding.
    \item We release \textbf{\textsc{KITE}}, a real-world time series event detection benchmark with expert descriptions and annotations.
\end{itemize}

\section{Problem Formulation}
\label{sec:problem}

The objective is to detect all event instances in a multivariate time series that match a set of linguistic event descriptions.

\paragraph{Input.}
The input consists of a multivariate time series $\mathbf{X}\in\mathbb{R}^{T\times C}$ and a description set $\mathcal{L}=\{\ell_e\}_{e\in\mathcal{E}}$, where $T$ is the number of time steps, $C$ is the number of channels, $\mathcal{E}$ is the event type set, and $\ell_e$ describes the signal pattern of event type $e$.

\paragraph{Ground Truth.}
The ground truth is a set of event instances $\hat{\mathcal{Y}}=\{\hat{y}_m\}_{m=1}^{M}$. Each instance is $\hat{y}_m=(\hat{I}^{(m)},\hat{e}^{(m)})$, where $\hat{I}^{(m)}=[\hat{t}_{\mathrm{on}}^{(m)},\hat{t}_{\mathrm{off}}^{(m)}]$ is the temporal interval and $\hat{e}^{(m)}\in\mathcal{E}$ is the event type.

\paragraph{Output.}
A detector $f_\theta$ predicts a set of candidate event instances $\mathcal{Y}=f_\theta(\mathbf{X},\mathcal{L})=\{y_k\}_{k=1}^{K}$. Each prediction is $y_k=(I^{(k)},e^{(k)})$, where $I^{(k)}=[t_{\mathrm{on}}^{(k)},t_{\mathrm{off}}^{(k)}]$ is the predicted interval and $e^{(k)}\in\mathcal{E}$ is the predicted event type.

\paragraph{Objective.}
Predictions are evaluated by matching them to ground-truth instances. A valid match must identify the correct event type, i.e., $e^{(i)}=\hat{e}^{(j)}$, and align with the corresponding ground-truth interval. Temporal alignment can be measured by interval IoU or boundary deviation. Thus, the task requires both event identification, determining \textit{what} event occurs, and temporal localization, determining \textit{when} it occurs.

\section{Event Logic Tree}
\label{sec:elt}

\begin{table*}[t]
    \centering
    \scalebox{0.82}{
    \begin{tabular}{cccc}
        \toprule
        \textbf{Operator} & \textbf{Natural Language Description} & \textbf{Formalization} & \textbf{TS Instance} \\
        \midrule
        
        \textbf{\texttt{SEQ}} & 
        \makecell{\textit{``A \textcolor{colorA}{spike in A} is \textbf{followed by} a \textcolor{colorB}{drop in B}.''}\\\textit{``A \textcolor{colorB}{drop in B} occurs \textbf{after} a \textcolor{colorA}{spike in A}.''}} &
        ${\text{\texttt{SEQ}}}(\textcolor{colorA}{\text{Spike}_A}, \textcolor{colorB}{\text{Drop}_B})$ & 
        \includegraphics[height=1.4cm, keepaspectratio, valign=m]{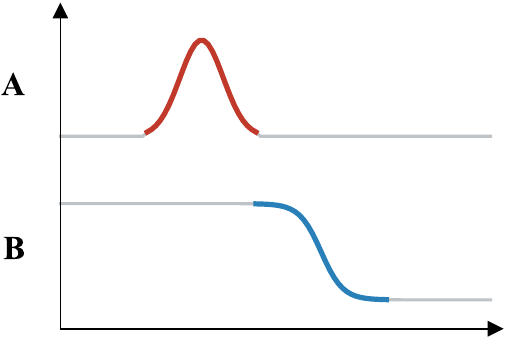} \\
        
        \textbf{\texttt{SYNC}} & 
        \makecell{\textit{``A \textcolor{colorA}{square wave in A} is \textbf{synchronized with} a \textcolor{colorB}{spike in B}.''}\\\textit{``A \textcolor{colorA}{square wave in A} and a \textcolor{colorB}{spike in B} occur \textbf{simultaneously}.''}} & 
        ${\text{\texttt{SYNC}}}(\textcolor{colorA}{\text{SquareWave}_A}, \textcolor{colorB}{\text{Spike}_B})$ & 
        \includegraphics[height=1.4cm, keepaspectratio, valign=m]{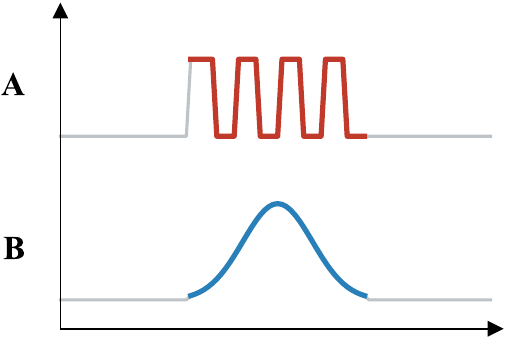} \\
        
        \textbf{\texttt{GUARD}} & 
            \makecell{\textit{``A \textcolor{colorA}{drop in A} is found \textbf{within} a \textcolor{colorB}{rise in B}.''}\\\textit{``A \textcolor{colorB}{rise in B} \textbf{encompasses} a \textcolor{colorA}{drop in A}.''}} & 
        ${\texttt{GUARD}}(\textcolor{colorA}{\text{Drop}_A}, \textcolor{colorB}{\text{Rise}_B})$ & 
        \includegraphics[height=1.4cm, keepaspectratio, valign=m]{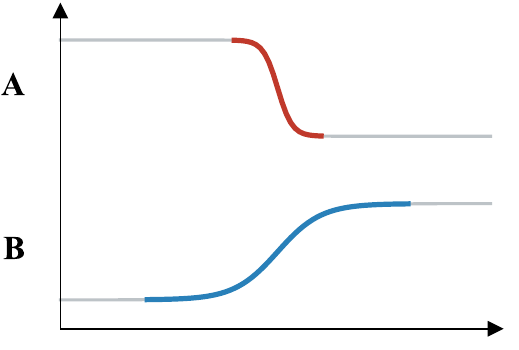} \\
        
        \textbf{\texttt{OR}} & 
        \makecell{\textit{``\textbf{Either} a \textcolor{colorA}{spike in A} \textbf{or} a \textcolor{colorB}{drop in B}.''}\\\textit{``\textbf{At least one} of a \textcolor{colorA}{spike in A} \textbf{or} a \textcolor{colorB}{drop in B} occurs.''}} & 
        ${\texttt{OR}}(\textcolor{colorA}{\text{Spike}_A}, \textcolor{colorB}{\text{Drop}_B})$ & 
        \includegraphics[height=1.4cm, keepaspectratio, valign=m]{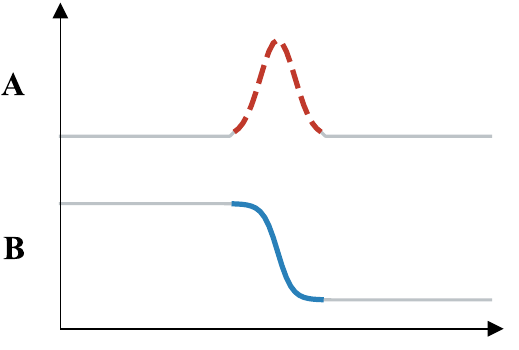} \\
        \bottomrule
    \end{tabular}
    }
        \caption{Core operators of Event Logic Tree.}
        \label{tab:operators}
\end{table*}

To achieve accurate event detection with faithful explanations, we need to explicitly model the internal structure of events. Based on the example in Section~\ref{sec:intro}, we define three desiderata for such a logic structure:

\paragraph{D1. Hierarchical Representation.}
The representation should capture how atomic patterns recursively form sub-patterns and finally the global event.

\paragraph{D2. Semantic Quantification.}
The representation should quantify how well a signal morphology matches its linguistic semantics.

\paragraph{D3. Topological Elasticity.}
The representation should define events by their temporal-logic structure rather than absolute duration, allowing temporal stretching or compression while preserving logical validity.

\paragraph{}To meet these requirements, we introduce \textbf{Event Logic Tree} (ELT), a tree-structured representation that separates an event's \underline{schema} from its \underline{instantiation}. The schema is parsed from the textual description only and specifies the event logic independently of any concrete time series sample. Each node is either a primitive or a composite:
\begin{equation}
\begin{aligned}
n_p &= \langle \tau, c \rangle, \\
n_\phi &= \langle [n_1,\ldots,n_k], op \rangle, \\
op &\in \{\texttt{SEQ}, \texttt{SYNC}, \texttt{GUARD}, \texttt{OR}\}.
\end{aligned}
\end{equation}
A primitive node $n_p$ denotes an atomic morphology $\tau$ on a physical channel $c$, such as ``\textit{a sharp pressure increase on the Pressure channel}''. A composite node $n_\phi$ recursively combines child nodes through a temporal-logic operator. Inspired by Allen's interval algebra~\citep{allen-1983-main-knowledge-temp}, our operator set is adapted to the grounding need: \texttt{SEQ} models succession, \texttt{SYNC} synchronization, \texttt{GUARD} containment, and \texttt{OR} alternative descriptions of the same event. Examples are shown in Table~\ref{tab:operators}, and the full schema and operator semantics are given in Appendix~\ref{app:elt_schema}. Figure \ref{fig:elt_ex} displays how the example description in Section \ref{sec:intro} is parsed into an event logic tree schema.

\begin{figure}[ht]
\centering
\scalebox{0.98}{
\includegraphics[width=\columnwidth]{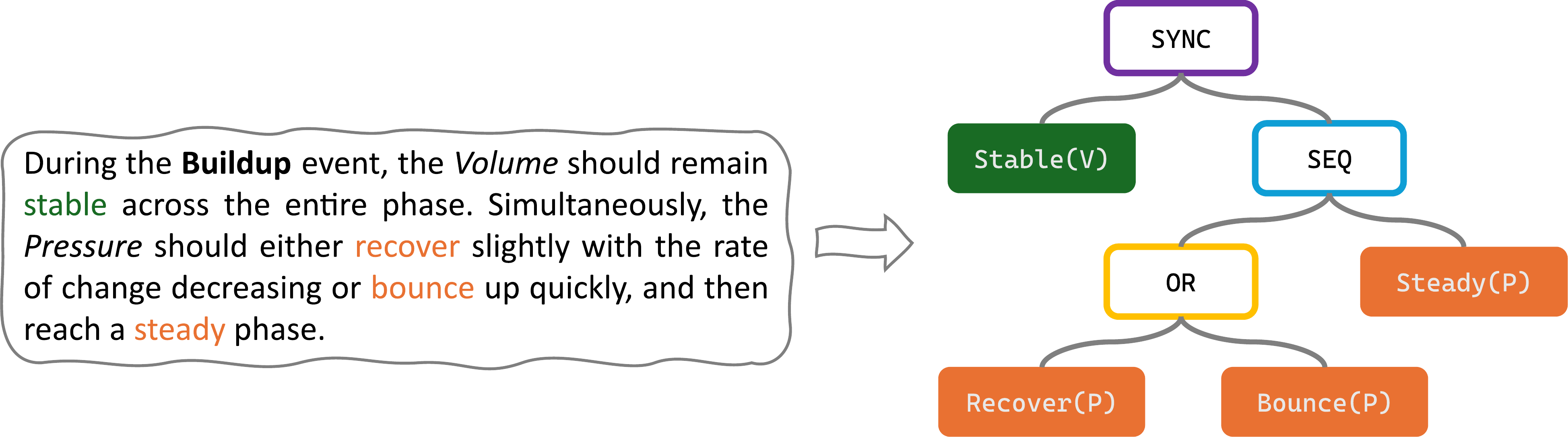}
}
\caption{An example of ELT schema parsing.}
\label{fig:elt_ex}
\end{figure}

Given a time series $X$, ELT is instantiated by grounding schema nodes to actual intervals. A primitive instance is defined as
\begin{equation}
\begin{aligned}
\hat n_p &= \langle n_p, I, \mu \rangle, \\
I &= [t_{\mathrm{on}}, t_{\mathrm{off}}], \\
\mu &= M_\tau(X, I, c) \in [0,1],
\end{aligned}
\end{equation}
where $\mu$ measures the semantic coherence between the segment $X[I,c]$ and the predicate $\tau$. Composite instances are computed bottom-up: their intervals span the intervals of their children, and their coherence scores are recursively aggregated by the corresponding operators. Thus, the root node represents the final event instance, with a numeric score reflecting both primitive-level grounding and global temporal consistency. Formal instantiation rules are provided in Appendix~\ref{app:elt_inst}.

To avoid physically invalid trees, ELT follows three validity principles: \textit{constructive composition}, requiring each composite to add non-trivial structure; \textit{temporal compactness}, preventing unbounded unexplained gaps inside a composite event; and \textit{physical exclusivity}, preventing one channel from supporting incompatible primitive states at the same time. These principles constrain grounding while preserving temporal elasticity; their formal definitions are given in Appendix~\ref{app:elt_axiom}.

Overall, ELT turns event detection into the construction of an instantiated proof tree: primitives provide signal-level evidence, operators organize the evidence into event-level logic, and confidence propagates from leaves to the root. We compare ELT with existing time series symbolic representations in Section~\ref{sec:tssr}.

\begin{figure*}[ht]
\centering
\scalebox{0.98}{
\includegraphics[width=\linewidth]{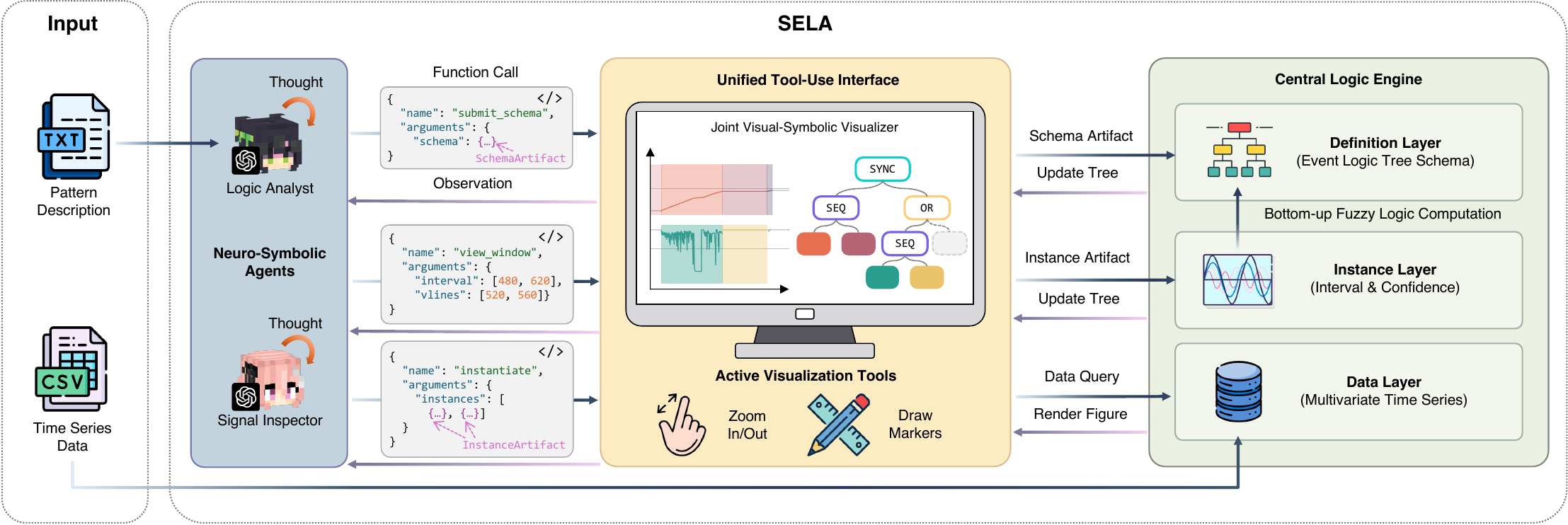}
}
\caption{Overview of the \textsc{SELA} system.}
\label{fig:SELA}
\end{figure*}

\section{The \textsc{SELA} Multi-Agent System}
\label{sec:sela}

We build \textsc{SELA}, a neuro-symbolic VLM agent system that uses ELT as the intermediate representation for language-guided time series event detection. The architecture is shown in Figure~\ref{fig:SELA}. We explain the core concepts as follows.

\paragraph{ELT-guided execution.}
\textsc{SELA} directly builds on the ELT representation introduced in Section~\ref{sec:elt}. \cww{The system first parses textual descriptions into logic trees. During grounding, \textsc{SELA} assigns intervals and semantic coherence scores to primitive nodes. These scores are then propagated bottom-up through the composite operator nodes, and the root node produces the final event interval and root confidence. In this way, the predictions are not just event intervals,} but instantiated proof trees whose root summarizes both primitive-level semantic evidence and global temporal consistency.

\paragraph{Agent workflow.}
\textsc{SELA} decomposes the task into two role-specific agents. The \textit{Logic Analyst} receives only the textual event description and submits an ELT schema. It does not access the time series, which prevents schema construction from being biased toward a particular sample. The \textit{Signal Inspector} receives the compiled schema and grounds it in the observed time series. It uses visualization tools to inspect global and local signal views, places optional temporal markers, and submits primitive instances with intervals and confidence scores \cww{that represent the level of semantic coherence under the agent's point of view.} The backend updates the ELT status for each agent submission, recomputes all composite scores, and returns a joint visualization of the signal evidence and the current logic-tree state. Agent prompts are given in Appendix~\ref{app:imp_details}.

\paragraph{Logic engine and confidence propagation.}
The \textsc{SELA} backend maintains three coupled states: the multivariate time series, the ELT schema, and the current primitive instantiations. Once the Signal Inspector submits primitive candidates, the logic engine recursively constructs the instantiated proof tree. Invalid compositions are filtered by hard gates, such as temporal-order violations or physical collisions on the same channel, while valid compositions receive confidence scores through the corresponding fuzzy operators. This makes any grounding failures explicit: a low root confidence can be traced back to weak primitive evidence, poor temporal alignment, missing branches, or violated operator constraints. 

When multiple mutually exclusive event candidates compete for the same record, we compare their normalized root confidence rather than the raw root confidence. This avoids unfairly penalizing a semantically richer tree only because it contains more conjunctive evidence. For an instantiated tree \(\hat T\), we use
\begin{equation}
\begin{aligned}
\mu_{\mathrm{norm}}(\hat T) &= \mu_{\mathrm{root}}(\hat T)^{1/K(\hat T)}, \\
K(\hat T) &= |\mathcal{L}^{*}(\hat T)|,
\end{aligned}
\end{equation}
where \(\mathcal{L}^{*}(\hat T)\) denotes the active primitive leaves of the instantiated tree.

\paragraph{Error control with artifacts.}
Agents interact with the backend through structured function calls and typed artifacts \citep{scale-agent}. The Logic Analyst submits a schema artifact, while the Signal Inspector submits an instantiation artifact containing primitive aliases, intervals, and confidence scores. Before updating the backend state, \textsc{SELA} validates the schema, parses artifacts, and checks action outputs against predefined formats. 

\section{The \textsc{KITE} Dataset Collection}
\label{sec:dataset}

We introduce the \textsc{KITE} (\underline{K}nowledge-\underline{I}nfused \underline{T}ime Series \underline{E}vents) benchmark, which is curated from two real-world domains: energy and climate.

\subsection{Data Collection}

\cww{\textsc{KITE} comprises \cam{two} sub-datasets:}

\paragraph{1. Pressure Test:}\cww{Pressure Test is a critical well exploration operation in the oil \& gas industry. The objective is to acquire underground pressure at different depths to characterize the reservoir. During pressure test jobs, sensors are installed at certain depths in drilling wells for remote measurements. We collected a dataset of 62 multivariate time series samples from 41 real underground pressure test jobs in drilling wells in the North Sea.} \cam{Pressure Test has two events: \textit{valid test} and \textit{lost seal}. Each time series sample has two channels: \textit{volume} and \textit{pressure}.}

\paragraph{2. Climate Hazard:}\cww{We created the Climate Hazard dataset by aligning two public datasets: the NOAA Integrated Surface Database (ISD)\footnote{\url{https://www.ncei.noaa.gov/data/global-hourly/access/}} and the NWS Storm Events database\footnote{\url{https://www.ncei.noaa.gov/stormevents/}}. Through manual filtering and calibration, we acquired \cam{86} multivariate time series samples with high-quality event labels.} \cam{We selected three hazard event types with support from accessible, high-quality knowledge as labels: \textit{dense fog}, \textit{frost freeze} and \textit{extreme cold}. Each climate time series sample has seven channels: \textit{visibility}, \textit{dew-point spread}, \textit{air temperature}, \textit{dew point}, \textit{wind chill}, \textit{wind speed}, and \textit{sea-level pressure}.}

\subsection{Data Annotation}
\label{sec:anno}

\cww{In general, the mission of data annotation is to obtain 1) the event labels (classes and boundaries), and 2) the event descriptions in natural language.}

For Pressure Test, the collected data were annotated in two stages. First, two field engineers with over 20 years of experience in the oil \& gas industry manually inspected every sample \tj{using Insitu Pro, a professional time series annotation software \citep{insitu-pro}}, to identify the classes and boundaries of all events. Their results were then cross-verified to \cww{resolve boundary disagreements}. In the second stage, the field engineers documented textual guidance for event detection, assisted by an experienced AI engineer.

\cww{For Climate Hazard, we first took the event boundaries from the NWS Storm Events database. However, the original timestamps reflect zone-level conditions rather than point-station measurements, which can introduce timing offsets at the individual station level, causing minor misalignments with the NOAA database. To solve this, we aggregated and summarized the official hazard event knowledge\footnote{\url{https://www.weather.gov/media/ajk/brochures/Wind_Chill_Temperature_Index.pdf}}\footnote{\url{https://www.weather.gov/media/directives/010_pdfs_archived/pd01016005c.pdf}} as textual descriptions. We had two annotators with climate science background manually inspect the signal samples to filter out abnormal data and calibrate the remaining event boundaries to align with the event descriptions. The results were cross-verified between the two annotators to resolve disagreements. Our full pipeline of constructing climate data is detailed in Appendix \ref{app:noaa}.}

\begin{table}[t]
\centering
    \scalebox{0.7}{
\begin{tabular}{llrrrrrr}
\toprule
 & & & \multicolumn{5}{c}{Length Statistics} \\
\cmidrule(lr){4-8}
Set &  & Ratio & Min & Q1 & Median & Q3 & Max \\
\midrule
\multirow{3}{*}{\shortstack{\textbf{PT}}} & \textit{TS Sample} &  & \cam{467} & \cam{944} & \cam{1313} & \cam{2228} & \cam{14145}\\
 & lost seal & 22.6 & \cam{151} & \cam{247} & \cam{346} & \cam{540} & \cam{1945}\\
 & valid test & 77.4 & \cam{262} & \cam{489} & \cam{800} & \cam{1363} & \cam{3386}\\
\midrule
\multirow{4}{*}{\shortstack{\textbf{CH}}} & \textit{TS Sample} &  & \cam{385} & \cam{661} & \cam{678} & \cam{1093} & \cam{1621}\\
 & dense fog & \cam{65.1} & \cam{13} & \cam{46} & \cam{76} & \cam{106} & \cam{293}\\
 & extreme cold & \cam{25.6} & \cam{228} & \cam{259} & \cam{384} & \cam{453} & \cam{756}\\
 & \cam{frost freeze} & \cam{9.3} & \cam{96} & \cam{120} & \cam{161} & \cam{396} & \cam{1152}\\
\bottomrule
\end{tabular}
}
\caption{Statistics by Dataset: \textbf{PT} = Pressure Test, \textbf{CH} = Climate Hazard.
Lengths are event durations in samples; \textit{TS Sample} is the series length.
Ratio is the percentage of samples containing that class.}
    \label{tab:dataset_stats}
\end{table}

\subsection{Dataset Statistics}

\cww{We provide the statistical details of the \cam{two} \textsc{KITE} datasets in Table \ref{tab:dataset_stats}. Overall, the time series samples are long, especially those in \cam{Pressure Test}, which have a median length of 1313 timestamps. For Pressure Test and Climate Hazard, the label distributions are imbalanced, posing additional challenges for supervised learning. Moreover, both sample lengths and event durations vary substantially across the \cam{two} datasets, requiring models to understand event semantics whose concrete signal realizations can be highly variable. This observation empirically supports the desideratum of \textit{Topological Elasticity} in Section \ref{sec:elt}.
}

\section{Experiments}
\label{sec:exp}

\cww{We develop our analysis based on two research questions:}

\paragraph{RQ 1.} \cww{How well do LLMs/VLMs understand event structures from natural language descriptions?}

\paragraph{RQ 2.} \cww{Could Event Logic Tree improve VLM agents' performance in time series event detection?}


\subsection{RQ 1. Benchmarking ELT Parsing}

\paragraph{Experimental Setup.}
We construct ground-truth ELTs for \cam{five} events across the two \textsc{KITE} datasets. We evaluate \cam{four} models: GPT-5\footnote{\url{https://ai.azure.com/catalog/models/gpt-5}} \citep{gpt5}, GPT-4.1\footnote{\url{https://ai.azure.com/catalog/models/gpt-4.1}} \citep{gpt4.1}, Qwen-3.5 27B\footnote{\url{https://huggingface.co/Qwen/Qwen3.5-27B}} \citep{qwen3.5}, and Gemma-4 31B\footnote{\url{https://huggingface.co/google/gemma-4-31B-it}} \citep{gemma4}. Each model parses each event five times, yielding \cam{100} parse runs in total.


\paragraph{Evaluation Dimensions.}
\yf{
We first report the \textit{pass rate} (Pass), the probability that a model produces a valid schema that can pass the syntax check within a fixed number of turns (20). Each valid parse is then compared against the human-parsed schema along four dimensions: \textit{Structural Fidelity} (Struct), measuring whether the tree topology is logically equivalent to the ground truth; \textit{Primitive Coverage} (Cov), measuring whether all ground-truth primitives are included; \textit{Primitive Accuracy} (Acc), assessing whether each predicted primitive is grounded in the original description and penalising fabricated primitives; and \textit{Semantic Alignment} (Sem), measuring how well each primitive's description matches the corresponding part of the original event description. Scores are assigned on a 0--5 scale by two independent VLM judges, Claude Opus-4.6 and GPT-5.5, averaged and normalised to $[0, 1]$, and validated by human. Implementation details of judges are provided in Appendix~\ref{app:elt-parse-quality}.}

\begin{table}[h]
\setlength{\tabcolsep}{5pt}
\setlength{\aboverulesep}{0pt}
\setlength{\belowrulesep}{0pt}
\setlength{\extrarowheight}{2pt}
\noindent\makebox[\columnwidth][l]{%

\resizebox{\columnwidth}{!}{%
\begin{tabular}{@{}lccccc}
\toprule
\textbf{Model} & \textbf{Pass} & \textbf{Struct} & \textbf{Acc} & \textbf{Cov} & \textbf{Sem} \\
\midrule
GPT-5           & \scorecell{1.00} & \scorecell{0.90} & \scorecell{1.00} & \scorecell{1.00} & \scorecell{1.00} \\
GPT-4.1           & \scorecell{1.00} & \scorecell{0.86} & \scorecell{1.00} & \scorecell{0.98} & \scorecell{0.98} \\
Qwen-3.5 27B       & \scorecell{0.88} & \scorecell{0.92} & \scorecell{1.00} & \scorecell{1.00} & \scorecell{0.99} \\
Gemma-4 31B       & \scorecell{1.00} & \scorecell{0.70} & \scorecell{1.00} & \scorecell{0.96} & \scorecell{0.85} \\
\bottomrule
\end{tabular}}}
\caption{ELT parse quality scores averaged over five independent runs and normalised.}
\label{tab:tpq-results}
\end{table}


\begin{table}[ht]
  \centering
  \begingroup
  \renewcommand{\arraystretch}{0.98}
  \setlength{\tabcolsep}{4pt}
  \setlength{\extrarowheight}{0pt}
  \scalebox{0.76}{
  \begin{tabular}{clcccc}
    \toprule
    \multicolumn{2}{c}{\multirow{2}{*}{\textbf{Method}}} & \multicolumn{2}{c}{\textbf{Pressure Test}} & \multicolumn{2}{c}{\textbf{Climate Hazard}}\\
    \cmidrule(lr){3-4} \cmidrule(lr){5-6}
    & & \cam{\textbf{\small$\mathcal{M}$F1@0.5}} & \cam{\textbf{\small$\mathcal{M}$F1@0.9}} & \cam{\textbf{\small$\mathcal{M}$F1$^{\texttt{Cov}}$}} & \cam{\textbf{\small$\mathcal{M}$F0.5$^{\texttt{Cov}}$}}\\
    \midrule
    \multicolumn{2}{c}{Random} & \cam{9.0{\text{\small±4.2}}} & \cam{0.0{\text{\small±0.0}}} & \cam{12.0{\text{\small±0.6}}} & \cam{9.2{\text{\small±0.5}}} \\
    \midrule
    \multirow{6}{*}{\shortstack{Sup.\\Models}} 
      & CNN & \cam{38.6{\text{\small±2.3}}} & \cam{22.4{\text{\small±2.5}}} & \cam{45.1{\text{\small±3.2}}} & \cam{40.7{\text{\small±2.2}}} \\
      & Trans. & \cam{27.7{\text{\small±2.1}}} & \cam{17.7{\text{\small±1.1}}} & \cam{46.4{\text{\small±6.5}}} & \cam{40.8{\text{\small±6.6}}} \\
      & Timer & \cam{32.0{\text{\small±3.4}}} & \cam{3.3{\text{\small±0.4}}} & \cam{18.0{\text{\small±0.4}}} & \cam{17.1{\text{\small±0.6}}} \\
      & Moment & \cam{43.6{\text{\small±5.4}}} & \cam{3.6{\text{\small±1.2}}} & \cam{20.9{\text{\small±5.3}}} & \cam{20.4{\text{\small±5.6}}} \\
      & Chronos & \cam{43.4{\text{\small±5.6}}} & \cam{2.6{\text{\small±0.5}}} & \cam{37.7{\text{\small±6.8}}} & \cam{39.9{\text{\small±8.3}}} \\
      & \cam{TimesFM} & \cam{46.6{\text{\small±6.1}}} & \cam{3.9{\text{\small±0.6}}} & \cam{44.5{\text{\small±3.0}}} & \cam{47.4{\text{\small±3.4}}} \\
    \midrule
    \multirow{5}{*}{\shortstack{GPT-4.1}} 
      & Num. \textit{Zs} &  \cam{15.4{\text{\small±3.9}}}  &  \cam{12.6{\text{\small±2.9}}} &  \cam{38.3{\text{\small±2.6}}} &  \cam{46.2{\text{\small±2.4}}} \\
      & Num. \textit{Fs} & \cam{25.4{\text{\small±6.0}}}  &  \cam{17.2{\text{\small±5.0}}} & \cam{43.4{\text{\small±3.1}}} &  \cam{49.4{\text{\small±3.5}}} \\
      & VL-T. \textit{Zs} &  \cam{37.5{\text{\small±1.2}}} &  \cam{20.3{\text{\small±2.5}}} &  \cam{9.6{\text{\small±0.1}}} &  \cam{11.0{\text{\small±0.3}}} \\
      & VL-T. \textit{Fs} &  \cam{40.1{\text{\small±2.2}}} &  \cam{24.9{\text{\small±3.3}}} & \cam{12.8{\text{\small±1.3}}} & \cam{15.2{\text{\small±2.1}}} \\
      & \textbf{\textsc{SELA}} \textit{Zs} &  \cam{43.9{\text{\small±2.3}}} &  \cam{32.7{\text{\small±5.5}}} &  \textbf{\cam{52.2{\text{\small±5.2}}}} &  \cam{\underline{53.9{\text{\small±5.2}}}} \\
    \midrule
    \multirow{5}{*}{\shortstack{GPT-5 \\\textit{thinking}}} 
      & Num. \textit{Zs} & \cam{53.2{\text{\small±3.1}}} & \cam{49.9{\text{\small±4.1}}} & \cam{40.8{\text{\small±3.2}}} & \cam{45.7{\text{\small±3.6}}} \\
      & Num. \textit{Fs} & \cam{\underline{57.8{\text{\small±6.0}}}}  &  \cam{\underline{54.2{\text{\small±3.1}}}} & \cam{43.4{\text{\small±1.6}}} & \cam{49.4{\text{\small±1.1}}} \\
      & VL-T. \textit{Zs} & \cam{51.9{\text{\small±2.1}}} & \cam{42.8{\text{\small±3.5}}} & \cam{23.8{\text{\small±1.9}}} & \cam{26.0{\text{\small±3.3}}} \\
      & VL-T. \textit{Fs} & \cam{53.6{\text{\small±3.2}}} &  \cam{46.0{\text{\small±2.6}}} & \cam{30.0{\text{\small±3.0}}} & \cam{33.2{\text{\small±4.0}}} \\
      & \textbf{\textsc{SELA}} \textit{Zs} & \textbf{\cam{66.8{\text{\small±3.9}}}} & \textbf{\cam{60.0{\text{\small±3.6}}}} & \cam{\underline{50.7{\text{\small±1.5}}}} & \textbf{\cam{57.1{\text{\small±2.2}}}} \\
    \midrule
    \multicolumn{2}{c}{\textit{Human}} & \textit{82.3} & \textit{79.7} & \textit{69.5} & \textit{75.9}\\
    \bottomrule
  \end{tabular}
  }
  \caption{Comparison of different models and methods on the \cam{two} \textsc{KITE} datasets. \cam{Zs=Zero-shot; Fs=Few-shot In-Context Learning.}}
  \label{tab:main_result}
  \endgroup
\end{table}


\newcommand{\up}[1]{\textsuperscript{\scriptsize$\uparrow$#1}}
\newcommand{\down}[1]{\textsuperscript{\scriptsize$\downarrow$#1}}

\begin{table*}[t]
\centering
\setlength{\tabcolsep}{3.0pt}
\renewcommand{\arraystretch}{1.05}
\scalebox{0.85}{
\begin{tabular}{
@{}l
ccccc
@{\hspace{8pt}}
ccccc
@{}
}
\toprule

&
\multicolumn{5}{c}{\textbf{Pressure Test} ($\mathcal{M}$-F1)}
&
\multicolumn{5}{c}{\textbf{Climate Hazard} ($\mathcal{M}$-Coverage)}
\\
\cmidrule(lr){2-6}
\cmidrule(lr){7-11}

&
\multicolumn{2}{c}{\textbf{GPT-5}}
&
\multicolumn{2}{c}{\textbf{GPT-4.1}}
&
\textbf{Avg.}
&
\multicolumn{2}{c}{\textbf{GPT-5}}
&
\multicolumn{2}{c}{\textbf{GPT-4.1}}
&
\textbf{Avg.}
\\
\cmidrule(lr){2-3}
\cmidrule(lr){4-5}
\cmidrule(lr){7-8}
\cmidrule(lr){9-10}

\textbf{Variant}
& \textbf{@0.5}
& \textbf{@0.9}
& \textbf{@0.5}
& \textbf{@0.9}
&
& \textbf{F1}
& \textbf{F0.5}
& \textbf{F1}
& \textbf{F0.5}
&
\\
\midrule

\textbf{SELA}
& \underline{66.8}
& \underline{60.0}
& \underline{43.9}
& \underline{32.7}
& \textbf{50.9}
& \underline{50.7}
& \underline{57.1}
& \textbf{52.2}
& \textbf{53.9}
& \textbf{53.5}
\\

\multicolumn{11}{@{}l}{\itshape Event-logic structure} \\

\quad w/o ELT (Geo. Mean)
& 41.8\down{25.0}
& 39.1\down{20.9}
& 41.5\down{2.4}
& 32.2\down{0.5}
& 38.7\down{12.2}
& 29.4\down{21.3}
& 34.5\down{22.6}
& 48.5\down{3.7}
& 49.7\down{4.2}
& 40.5\down{13.0}
\\

\quad w/o ELT (Arith. Mean)
& 41.8\down{25.0}
& 39.2\down{20.8}
& 42.9\down{1.0}
& \underline{32.7}\up{0.0}
& 39.2\down{11.7}
& 29.6\down{21.1}
& 34.7\down{22.4}
& 47.4\down{4.8}
& 48.6\down{5.3}
& 40.1\down{13.4}
\\

\quad w/o ELT (Minimum)
& 34.7\down{32.1}
& 32.0\down{28.0}
& 38.1\down{5.8}
& 28.9\down{3.8}
& 33.4\down{17.5}
& 26.4\down{24.3}
& 31.1\down{26.0}
& 48.6\down{3.6}
& 50.1\down{3.8}
& 39.1\down{14.4}
\\

\multicolumn{11}{@{}l}{\itshape Confidence scoring} \\

\quad w/o $1/K$ norm
& \underline{66.8}\up{0.0}
& \underline{60.0}\up{0.0}
& 42.6\down{1.3}
& 32.3\down{0.4}
& \underline{50.4}\down{0.5}
& 49.4\down{1.3}
& 55.9\down{1.2}
& \underline{51.7}\down{0.5}
& \underline{53.2}\down{0.7}
& \underline{52.6}\down{0.9}
\\

\quad w/o operator penalties
& \textbf{67.2}\up{0.4}
& \textbf{61.8}\up{1.8}
& 42.8\down{1.1}
& 27.8\down{4.9}
& 49.9\down{1.0}
& \textbf{51.0}\up{0.3}
& \textbf{58.7}\up{1.6}
& 44.3\down{7.9}
& 44.8\down{9.1}
& 49.7\down{3.8}
\\

\multicolumn{11}{@{}l}{\itshape Inspection tool} \\

\quad w/o zoom
& 63.2\down{3.6}
& 51.8\down{8.2}
& \textbf{45.5}\up{1.6}
& \textbf{33.4}\up{0.7}
& 48.5\down{2.4}
& 49.9\down{0.8}
& 56.0\down{1.1}
& 51.4\down{0.8}
& 50.9\down{3.0}
& 52.1\down{1.4}
\\

\bottomrule
\end{tabular}
}

\caption{
\cam{
ELT ablations on Pressure Test and Climate Hazard.
Avg.\ averages the four backbone--metric entries.
\textbf{Bold} and \underline{underline} denote the best and second-best
results per column.}
}
\label{tab:elt-ablation}

\end{table*}

\paragraph{Results.}
\cam{As shown in Table~\ref{tab:tpq-results}, all valid parses achieve perfect Primitive Accuracy (1.00). GPT-5 attains perfect Coverage and Semantic Alignment, with strong Structural Fidelity (0.90). Qwen-3.5 27B shows the weakest Pass Rate (0.88) but achieves the highest Structural Fidelity (0.92) when successful. GPT-4.1 trails in Structural Fidelity (0.86), occasionally collapsing complex operator relations into simpler ones. Gemma-4 31B scores lowest on both Structural Fidelity (0.70) and Semantic Alignment (0.85), over-compressing primitive descriptions.}

\subsection{RQ 2. Evaluation on \textsc{KITE} Datasets}

\paragraph{Experimental Setup.} We compare \textsc{SELA} with \cam{four} groups of baselines: (i) \textbf{Random guessing}; (ii) \textbf{supervised models}, including CNN \citep{CNN}, Transformer \citep{transformer}, and fine-tuned time series foundation models, i.e., Timer \citep{liu-2024-timer}, Moment \citep{goswami-2024-moment}, Chronos \citep{ansari-2024-chronos}, \cam{and TimesFM \citep{timesfm}}; (iii) \textbf{few-shot/zero-shot LLM/VLM baselines}, including Numeric (\citealp{promptcast,llmtime}), which directly takes numerical time series as input, and VL-Time \citep{vl-time}, which lets VLMs reason over time series visualizations; (iv) \textbf{human}, where two data scientists \cww{with no prior task knowledge} independently identify events based only on the provided pattern descriptions.

\paragraph{Evaluation Metrics.} \cww{For Pressure Test, the event boundaries are definite and clear, as the signals are grounded in human engineer operations. Therefore, we} apply IoU-based \cam{macro} F1 scores (\cam{\textbf{$\mathcal{M}$F1@}}) as metrics to measure performance, following previous work on time series event detection in the oil \& gas domain \citep{khaouja-2025-do}. We set two IoU thresholds, 0.5 and 0.9, to better distinguish models by their ability to perform precise grounding. \cww{For Climate Hazard, event boundaries are more gradual. We therefore follow \citet{NEURIPS2018_8f468c87} and adopt coverage-based \cam{macro} F1 and F0.5 scores (\cam{\textbf{$\mathcal{M}$F1$^{\texttt{Cov}}$}},  \cam{\textbf{$\mathcal{M}$F0.5$^{\texttt{Cov}}$}}). More details on the metric design are provided in Appendix \ref{app:metrics}.}

\begin{figure*}[p]
    \centering

    \begin{subfigure}{\textwidth}
        \centering
        \includegraphics[width=\textwidth]{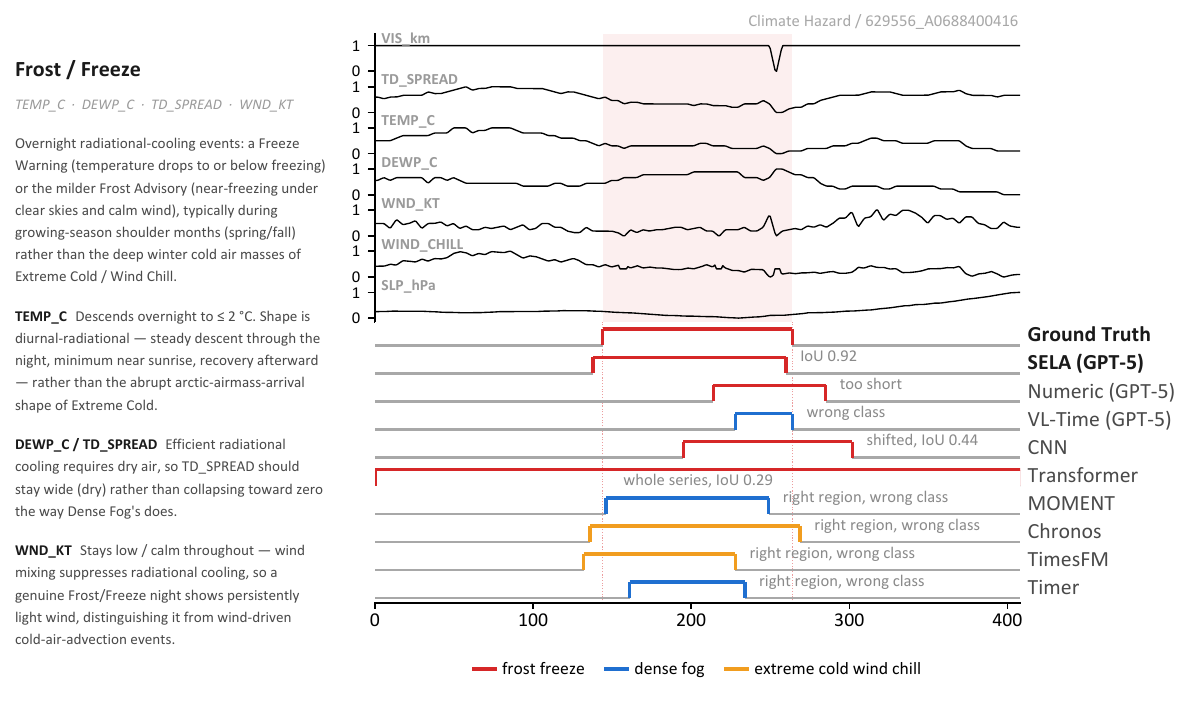}
        \caption{Comparison of event predictions on a Climate Hazard case.}
        \label{fig:case-cli}
    \end{subfigure}

    \vspace{0.5em}

    \begin{subfigure}{\textwidth}
        \centering
        \includegraphics[width=\textwidth]{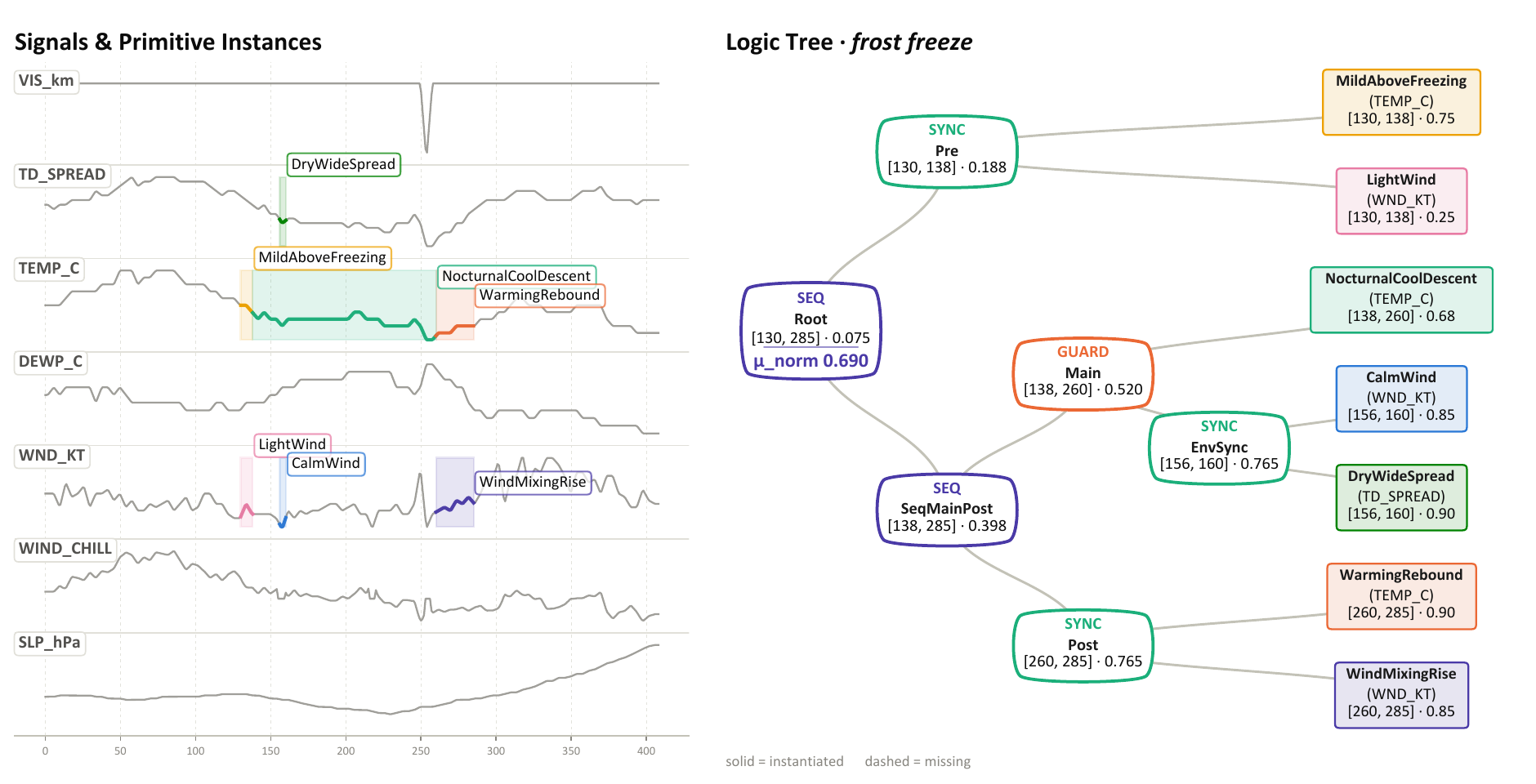}
        \caption{Instantiated primitive instances and Event Logic Tree for the selected case.}
        \label{fig:case-tree}
    \end{subfigure}

    \caption{Case study from Climate Hazard, showing predictions from different methods against the ground truth, together with the instantiated primitive instances and Event Logic Tree (ELT).}
    \label{fig:case-study-elt}
\end{figure*}

\paragraph{Implementation Details.}\cam{For CNN and Transformer, we add linear heads for timestamp-wise classification and fine-tune the entire models. For native time series foundation models, we freeze the model weights and fine-tune only the attached bounding-box classification heads.} All supervised models are tested using 5-fold cross-validation (80\% training, 20\% testing), and average scores \cam{over 3-seed runs} are reported. Regarding Numeric and VL-Time, we provide the prompts in Appendix \ref{app:baseline-prompts}. We also report the token usage of these VLM-based settings in Appendix \ref{app:token_budget}.

\paragraph{Results.} \cam{As shown in Table~\ref{tab:main_result}, \textsc{SELA} consistently outperforms zero-shot and few-shot Numeric and VL-Time baselines on both \textsc{KITE} datasets, despite itself operating in the zero-shot setting. The advantage is particularly clear for precise localization on Pressure Test: with GPT-5, \textsc{SELA} reaches 66.8 $\mathcal{M}$F1@0.5 and 60.0 $\mathcal{M}$F1@0.9, improving over few-shot VL-Time by 13.2 and 14.0 points, respectively. On Climate Hazard, \textsc{SELA} also achieves the strongest automated results, with GPT-4.1 obtaining the best $\mathcal{M}$F1$^{\texttt{Cov}}$ of 52.2 and GPT-5 obtaining the best $\mathcal{M}$F0.5$^{\texttt{Cov}}$ of 57.1. In contrast, supervised baselines show no consistent advantage in this low-resource setting; in particular, time series foundation models perform poorly under the strict $\mathcal{M}$F1@0.9 metric on Pressure Test, indicating limited boundary-localization ability. Overall, the consistent gains across GPT-4.1 and GPT-5 demonstrate that ELT-guided reasoning improves both event detection and precise temporal grounding, while human data scientists still retain a substantial performance margin.}

\paragraph{Ablation Study.} \cam{As shown in Table~\ref{tab:elt-ablation}, the event-logic structure is the most critical component of \textsc{SELA}. In the w/o ELT variants, we discard the tree structure and directly aggregate primitive-level confidence scores using three basic strategies: geometric mean, arithmetic mean, and minimum. Removing ELT causes substantial performance drops under all three aggregation strategies, reducing the average score by up to 17.5 points on Pressure Test and 14.4 points on Climate Hazard. In comparison, ablating confidence normalization, operator penalties, or the zoom inspection tool leads to smaller but consistent degradations. Among them, operator penalties are particularly important for Climate Hazard, while zoom mainly benefits precise localization on Pressure Test.}

\subsection{Case Study}
\cam{Figure~\ref{fig:case-study-elt} presents a \textit{Frost Freeze} case from Climate Hazard, where \textsc{SELA} closely matches the ground-truth. The instantiated ELT grounds the event in a sequential temperature evolution from mild-above-freezing conditions through sustained cooling to a warming rebound, while synchronizing and guarding these phases with complementary wind and dew-point-spread primitives. In contrast, Numeric predicts only a short fragment, VL-Time assigns the wrong event class, and CNN and Transformer substantially mislocalize the interval; the time series foundation models identify a relevant region but also predict the wrong class. This example illustrates how ELT-guided compositional reasoning helps \textsc{SELA} jointly resolve event semantics and temporal boundaries in noisy multivariate signals.}

\section{Related Work}

\subsection{Time Series Symbolic Representation}
\label{sec:tssr}

We analyze existing symbolic time series representation frameworks based on our desiderata (summarized in Table \ref{tab:desi}). 

\begin{table}[t]
  \centering
  \resizebox{\columnwidth}{!}{%
  \begin{tabular}{l c c c}
    \toprule
     & \makecell[c]{\textbf{D1: Hierarchical} \\ \textbf{Representation}} & \makecell[c]{\textbf{D2: Semantic} \\ \textbf{Quantification}} & \makecell[c]{\textbf{D3: Topological} \\ \textbf{Elasticity}} \\
    \midrule
    \makecell[l]{SAX / ABBA} 
    & \textcolor{red}{\ding{55}} & \textcolor{orange}{\large$\blacktriangle$} & \textcolor{red}{\ding{55}} \\
     
    \makecell[l]{Logical-Shapelets} 
    & \textcolor{orange}{\large$\blacktriangle$} & \textcolor{orange}{\large$\blacktriangle$} & \textcolor{ForestGreen}{\ding{51}} \\

    \makecell[l]{Z-Time} 
    & \textcolor{orange}{\large$\blacktriangle$} & \textcolor{orange}{\large$\blacktriangle$} & \textcolor{ForestGreen}{\ding{51}} \\

    \makecell[l]{Chronicle / STL} 
    & \textcolor{ForestGreen}{\ding{51}} & \textcolor{orange}{\large$\blacktriangle$} & \textcolor{red}{\ding{55}} \\
    \midrule
    \textbf{Event Logic Tree}
    & \textcolor{ForestGreen}{\ding{51}} & \textcolor{ForestGreen}{\ding{51}} & \textcolor{ForestGreen}{\ding{51}} \\
    \bottomrule
  \end{tabular}%
  }
  \caption{Comparison of frameworks against our desiderata. (\textcolor{ForestGreen}{\ding{51}}: Satisfied; \textcolor{orange}{$\blacktriangle$}: Limited / Partially satisfied; \textcolor{red}{\ding{55}}: Not satisfied)}
  \label{tab:desi}
\end{table}

(1) \cw{SAX (\citealp{sax-2003, sax}) and ABBA \citep{ABBA} map time series into string sequences where each character represents a specific amplitude. Moreover, the repeated sub-strings carry semantics but do not quantify coherence (partially satisfying D2), and do not allow hierarchical structure (failing D1) or complex topological relationships (failing D3).} 

(2) \cw{Logical-Shapelets \citep{logical-shapelets} represents entire time series with a series of discriminative sub-sequences , i.e., shapelets (satisfying D3). Z-Time \citep{z-time} improves the representation by making the length variable (satisfying D3). Both the shapelets and discrete representations of Z-Time carry limited semantic abstraction (partially satisfying D2). Regarding structural modeling, Logical-Shapelets only considers boolean relations, and while Z-Time uses Allen's algebra \citep{allen-1983-main-knowledge-temp}, only limited hierarchy structures can be represented by stacking temporal relation pairs. Therefore, both approaches partially satisfy D1.} 

(3) \cw{Chronicle System \citep{chronicle_rec_sys} applies graph structure to represent temporal events, where each sub-event is considered as a node in the graph, and STL \citep{stl} represents events as recursive logical formulas over signals intervals, both satisfying D1. However, both systems model semantic coherence as binary values (true/false) (partially satisfying D2), and their reliance on actual interval durations does not meet D3.}



\cw{In summary, existing symbolic systems can only make structured representations from time series but not language, thus not satisfying the requirements of L-TSED. In contrast, our proposed ELT framework uses tree structures based on Allen's algebra and boolean operators to support D1. The definition of tree schema and instantiation on actual time series data are separated to satisfy D3. ELT supports quantifying semantic coherence with neural models over any combination of basic time series signal attributes, satisfying D2.}

\subsection{Multimodal LLMs for TS Reasoning}

Although time-series foundation models show promising zero-shot or limited-supervision capabilities (\citealp{ansari-2024-chronos, liu-2024-timer, goswami-2024-moment, timesfm}), they are not natively designed for language-conditioned event localization and generally require task-specific adaptation for downstream detection tasks. Recently, LLMs' potential to serve as few/zero-shot time series reasoners has been widely explored. ChatTS \citep{chatts} designed an encoder with synthetic data to align time series with language for QA tasks. However, the domain gap between synthetic and real-world data is difficult to bridge, and embedding time series can still introduce the risk of hallucination. VL-Time \citep{vl-time} visualizes time series data as figures to directly adopt vision-language reasoning capabilities without extra TS-language alignment. While it has shown advantages in classification, visualization inevitably loses precision, and hallucination is still a risk. Our approach employs active visualization tools to overcome precision loss and effectively mitigates hallucination with ELT representations.

\section{Conclusion}

We introduce a new language-guided setting for time-series event detection and present \textsc{KITE}, the first real-world benchmark for this task. To address the low-resource and knowledge-intensive nature of this setting, we propose the Event Logic Tree and build \textsc{SELA}, a zero-shot neuro-symbolic agent system that combines ELTs with large vision-language models. Experiments, ablations, and case studies show that ELT-guided reasoning improves precise event localization and provides interpretable evidence for human-in-the-loop verification.



\section*{Limitations}

\paragraph{Benchmark scope.}
\cam{\textsc{KITE} remains limited in scale and domain coverage because expert event descriptions and interval annotations are costly to obtain. We plan to include more real-world multivariate time-series data from more key domains in the future.}

\paragraph{Faithfulness of instantiation.}
\cam{Currently, we rely on VLMs to provide confidence scores for primitive-level text--time-series alignment during instantiation. Human evaluation of such grounding faithfulness is extremely costly since it requires domain experts to evaluate the alignment between all primitive-level textual descriptions and detected signal sections. Direct automatic evaluation is also difficult because reliable embedding models specifically designed to align domain-specific time-series primitives with textual descriptions are currently lacking. We therefore plan to reduce the impact of LLM/VLM subjectivity by developing a foundation model that can more objectively evaluate the alignment between domain-specific primitive-level textual descriptions and detected signal segments.}

\paragraph{System complexity.}
\cam{\textsc{SELA} relies on LLM/VLM reasoning, which is costly and inherently stochastic, and may lead to variation in parsing, grounding, and confidence estimation across runs. This randomness is difficult to remove completely in an agent-based system. Future work will explore more stable and efficient models, together with improved reasoning and grounding methods, to reduce computational cost and support low-latency inference. We will also maintain the goal of transparency and explainability throughout the reasoning process, so that predictions and intermediate evidence can be inspected by domain experts and safely used under human supervision.}

\paragraph{Potential risks.} Potential risks arise if incorrect event detections or overconfident explanations are used for operational decisions in safety-sensitive settings. \textsc{SELA} is intended as a research and human-in-the-loop decision-support framework rather than an autonomous safety-critical system, and its predictions should be verified by domain experts.

\section*{Acknowledgments}

This work contains information provided by the North Sea Transition Authority. 

\cam{We thank the members of the SLB AI Lab for their support and valuable feedback throughout this work. We are especially grateful to Josselin Kherroubi, Director of the SLB AI Lab, for providing the resources that made this research possible. We also thank the anonymous reviewers, particularly Reviewer 2qFd, for their thoughtful comments and constructive suggestions, which helped improve the paper. Finally, we thank Prof. Matthieu Labeau and Prof. Fabian Suchanek from Télécom Paris for their valuable discussions and insightful advice.}


\bibliography{custom}

\clearpage
\appendix

\startcontents[appendix]   

\vspace{0.3\baselineskip}

\noindent{\fontsize{12}{8}\selectfont\textsc{Table of Contents}}

\vspace{0.03\baselineskip}
\noindent\rule{\columnwidth}{0.9pt}

\printcontents[appendix]{l}{1}{\setcounter{tocdepth}{2}}

\vspace{0.2\baselineskip}
\noindent\rule{\columnwidth}{0.9pt}

\section{Event Logic Tree}
\label{app:elt}

\subsection{ELT Schema}
\label{app:elt_schema}

\cw{The Event Logic Tree schema is parsed from textual descriptions only, independent of actual time series data. One event corresponds to one tree.}

\begin{definition}[Node Schema]
    The Event Logic Tree Schema is denoted as $\mathcal{S}$. Any node in the schema $n\in\mathcal{S}$ belongs to two categories: primitives and composites.
    \paragraph{1. Primitive (Leaf Node)}: The atomic signal patterns over single physical channels, denoted as $n_p=\langle \tau, c \rangle$, where:
    \begin{itemize}
        \item $\tau$: The \textbf{semantic predicate} that describes signal morphology, which can be a compound of basic time series attributes, e.g., "a steep linear increase with high-frequency noise".
        \item $c$: The \textbf{physical channel} index where this atomic pattern would occur.
    \end{itemize}

    \paragraph{2. Composite (Internal Node).} The temporal-logic relations that form the global event hierarchically with the primitives, denoted as $n_{\phi}=\langle \mathcal{N}, op \rangle$, where:
    \begin{itemize}
        \item $\mathcal{N}=[n_1,\dots,n_k]$: An ordered list of child nodes, either primitives or lower-level composites.
        \item $op\in \{\texttt{SEQ}, \texttt{SYNC}, \texttt{GUARD}, \texttt{OR}\}$: The \textbf{temporal-logic operator} defining the relationship between child nodes.
    \end{itemize}
\end{definition}

\subsection{Constitutive Axioms}
\label{app:elt_axiom}

\cw{Based on the definitions above, we establish three axioms to serve as \tj{preconditions} for valid schema construction, and properly control the search space during instantiation.}

\begin{axiom}[Constructive Composition]
\label{ax:structural_gain}
\cw{A composite node must add structural meaning. Formally, for any composite $n_\phi$, $|\mathcal{N}| \ge 2$.}
\end{axiom}

\paragraph{Remark.} Axiom 1 excludes bad structures with self-nesting, and any logical \texttt{NOT} as composite with a single child. We require all primitives to be defined \textit{positively}.

\begin{axiom}[Temporal Compactness]
\label{ax:semantic_contiguity}
\cw{The semantics of any node must cover its full temporal span. Any undefined gap inside the temporal spans of composites must be bounded by a hyperparameter.} 
\end{axiom}

\paragraph{Remark.} Primitives are considered to be compact by definition. For any composite, assuming the compactness tolerance is 1, a composite whose children are assigned disjoint intervals like $[1,2]$ and $[4,6]$ will be prohibited by Axiom 2.

\begin{axiom}[Physical Exclusivity]
\label{ax:atomicity}
\cw{A physical channel $c$ can only support one active primitive state at any time point $t$.}
\end{axiom}

\subsection{Instantiation of ELT}
\label{app:elt_inst}

\begin{definition}[Node Instance]
\label{def:node}
An instance implies designating a specific time interval to a schema node $n$. 


\paragraph{1. Primitive Instance.}
A primitive instance $\hat{n}_p$ is defined as: $\langle n_p, \mathcal{I}, \mu \rangle$, where:

\begin{itemize}
    \item $\mathcal{I}= [t_{\texttt{on}}, t_{\texttt{off}}]$: The detected time \textbf{interval}.
    \item $\mu = \mathcal{M}_{\tau}(X,\mathcal I,c)\in[0,1]$: The \textbf{semantic coherence score}, which quantifies the alignment between the signal and the predicate. \tj{The implementation of the semantic function $\mathcal{M}_{\tau}$ is not restricted to rule-based functions. Measurement can be evaluated based on the distance between the embeddings of linguistic description and signal pattern in an aligned latent space.}
    
\end{itemize}

\paragraph{2. Composite Instance.}
A composite instance $\hat{n}_{\phi}$ is defined as $\langle n_{\phi}, \mathcal{I}_{\phi}, \mu_{\phi} \rangle$:

\begin{itemize}
    \item $\mathcal{I}_{\phi} = [\min_{\hat{n} \in \hat{\mathcal{N}}} t_{\texttt{on}}^{(\hat{n})}, \max_{\hat{n} \in \hat{\mathcal{N}}} t_{\texttt{off}}^{(\hat{n})}]$: The \textbf{composite interval}, which is the temporal span of constituents' intervals (subject to Axiom \ref{ax:semantic_contiguity}). $\hat{\mathcal{N}}=[\hat n_1,\ldots,\hat n_k]$ denotes the child instances.
    \item $\mu_{\phi} = op(\hat{n}_1, \dots, \hat{n}_k)$: The \textbf{recursive confidence score} by applying the operator on the constituents.
\end{itemize}
\end{definition}

\tj{To ensure the coherence of Axiom \ref{ax:atomicity}, we need to introduce a collision operator $\Psi$. If the Axiom \ref{ax:atomicity} is violated by any instance, $\Psi=1$.
\begin{definition}[Channel-Semantic Collision]
\label{def:collision}
\begin{equation}
\begin{aligned}
\Psi(\hat n_A,\hat n_B)
&=
\max_{\substack{p \in \Pi(\hat n_A)\\ q \in \Pi(\hat n_B)}}
\mathbf{1}\Big[
\big(c_p=c_q\big)\ \wedge\ \\&\big(\ell(\mathcal I_p \cap \mathcal I_q)>\epsilon\big)
\Big]
\end{aligned}
\label{eq:collision}
\end{equation}
\end{definition}
$\Pi(\hat{n})$ denotes the set of \textbf{primitive} instances descendants of $\hat{n}$, if $\hat{n}$ is a composite instance, or the set of $\hat{n}$ itself, if $\hat{n}$ is a primitive instance. $\mathbf{1}[\cdot]$ denotes the indicator function.
For discrete sampling, a change point may belong to both adjacent intervals at the same time. This violation case should be ignored. Therefore, Semantic Nullity Threshold $\epsilon$ is introduced to set up a tolerance.
}


\begin{definition}[Temporal-Logic Operators]
    We define the core operator set to comprise four operators: $\mathcal{O}_{\text{core}} = \{\texttt{SEQ}, \texttt{SYNC}, \texttt{GUARD}, \texttt{OR}\}$ \tj{(Table \ref{tab:operators})}. The operators aggregate confidence scores in a bottom-up manner with product T-norm, while respecting Temporal Compactness and Physical Exclusivity.


\paragraph{1. \texttt{SEQ} (Sequence).}
\texttt{SEQ} represents the temporal precedence relationship (B follows A). The confidence score is computed with product T-norm while controlled by validity gates:
\begin{align}
    {\texttt{SEQ}}(\hat{n}_A, \hat{n}_B) = &(\mu_A\cdot \mu_B) \cdot (1 - \Psi(\hat{n}_A, \hat{n}_B)) \nonumber \\ \cdot &\mathcal{G}_{\text{causal}}(\hat{n}_A, \hat{n}_B) \cdot \mathcal{G}_{\text{coh}}(\hat{n}_A, \hat{n}_B)
\end{align}
where the term $(1-\Psi(\hat{n}_A, \hat{n}_B))$ enforces Physical Exclusivity. The causality gate $\mathcal{G}_{\text{causal}}(\hat{n}_A, \hat{n}_B) = \mathbf{1}[t_{\texttt{on}}^B > t_{\texttt{on}}^A]\cdot \mathbf{1}[t_{\texttt{off}}^B > t_{\texttt{off}}^A]$ enforces the temporal precedence relationship. The coherence gate $\mathcal{G}_{\text{coh}}(\hat{n}_A, \hat{n}_B) = \mathbf{1}[(t_{\texttt{on}}^B - t_{\texttt{off}}^A - \delta) < 0]$ prohibits any significant semantic gap beyond the bound $\delta$ to ensure the Temporal Compactness principle.


\paragraph{2. \texttt{SYNC} (Synchronization).}
\texttt{SYNC} represents temporal identity ($\mathcal{I}_A = \mathcal{I}_B$). A penalty term based on IoU, \tj{denoted as $\mathrm{IoU}(\mathcal I_A,\mathcal I_B)=\frac{\ell(\mathcal I_A \cap \mathcal I_B)}{\ell(\mathcal I_A \cup \mathcal I_B)}$},  is applied to penalize the joint confidence in case of temporal misalignment.

\begin{align}
    \texttt{SYNC}(\hat{n}_A, \hat{n}_B) &= (\mu_A \cdot \mu_B) \cdot (1 - \Psi(\hat{n}_A, \hat{n}_B)) \nonumber \\
    &\cdot \exp\left( -\frac{1 - \text{IoU}(\mathcal{I}_A, \mathcal{I}_B)}{\kappa} \right)
\end{align}
where $\kappa$ is the alignment tolerance.

\paragraph{3. \texttt{GUARD} (Containment).}
\texttt{GUARD} represents $\mathcal{I}_A \subset \mathcal{I}_B$. Boundaries of $A$ that spill outside $B$ are penalized:
\begin{align}
    {\texttt{GUARD}}(\hat{n}_A, \hat{n}_B) &= (\mu_A\cdot \mu_B) \cdot (1 - \Psi(\hat{n}_A, \hat{n}_B)) \nonumber \\ 
    &\cdot \exp\left( -\frac{\Delta_{\texttt{on}} + \Delta_{\texttt{off}}}{\sigma} \right)
\end{align}
where $\Delta_{\texttt{on}} = \max(0, t_{\texttt{on}}^B - t_{\texttt{on}}^A)$ and $\Delta_{\texttt{off}} = \max(0, t_{\texttt{off}}^A - t_{\texttt{off}}^B)$ quantify how much $A$ extends beyond $B$, controlled by temperature $\sigma$.

\paragraph{4. \texttt{OR} (Disjunction).}
\texttt{OR} models semantic alternatives for the same instance. We encourage the two alternatives to refer to the same interval by using the same IoU-based alignment penalty as \texttt{SYNC}:
\begin{align}
    {\texttt{OR}}(\hat{n}_A, \hat{n}_B) &= \max(\mu_A, \mu_B) \nonumber
    \\&\cdot \exp\left( -\frac{1 - \text{IoU}(\mathcal{I}_A, \mathcal{I}_B)}{\kappa} \right)
\end{align}
\end{definition}

\subsection{Operational Completeness} 
\label{sec:unified_coverage}
From a logical perspective, \texttt{SEQ}, \texttt{SYNC}, and \texttt{GUARD} are \textit{conjunctive}, representing the Boolean \texttt{AND} (co-existing relation \citep{z-time}) projected onto temporal relationships. We can thus unify them as a generalized \textbf{Temporal Conjunction} $\texttt{AND}_{\mathcal K}$:
\begin{equation}
\begin{aligned}
\mathrm{AND}_{\mathcal K}(\hat n_A,\hat n_B)
&=
(\mu_A \cdot \mu_B) \cdot
\mathcal K(\hat n_A,\hat n_B) \\
&\quad \cdot \big(1-\Psi(\hat n_A,\hat n_B)\big)
\end{aligned}
\end{equation}

where $\mathcal K$ maps to \texttt{SEQ}/\texttt{SYNC}/\texttt{GUARD} by working as
corresponding soft/hard validity gates.

\begin{proposition}[\cam{Conditional} Completeness]
\label{prop:completeness}
\cam{Under the constraints of Axioms \ref{ax:structural_gain}-\ref{ax:atomicity}, }the operator set $\mathcal{O}_{core}$ forms a complete basis for the 13 fundamental Allen's interval relations.
\end{proposition}


\paragraph{Proof Sketch.}
We demonstrate that the generalized conjunction $\texttt{AND}_{\mathcal K}$ (as \texttt{SEQ, SYNC, GUARD}) exhaustively cover the Allen's algebra partition: 
\paragraph{(1) The Precedence Group (\texttt{SEQ}):} The relations implying temporal precedence are handled by the ${\texttt{SEQ}}$. The causality gate ($t_{\texttt{on}}^B > t_{\texttt{on}}^A \land t_{\texttt{off}}^B > t_{\texttt{off}}^A$) distinguishes it from ${\texttt{SYNC}}$ and ${\texttt{GUARD}}$.
(i) \textit{Meets} and \textit{Before}: While \textit{Meets} naturally corresponds to ${\texttt{SEQ}}$, the disjoint \textit{Before} relation is bounded by horizon $\delta$ to prevent any semantic gap that would violate Axiom \ref{ax:semantic_contiguity}. Therefore, the \textit{Before} relation is treated as a relaxed variant of \textit{Meets}.
(ii) \textit{Overlaps}: a special case corresponding to ${\texttt{SEQ}}$ if and only if Physical Exclusivity is ensured across all channels over the overlapping interval ($\Psi=0$). 

\paragraph{(2) The Containment Group (\texttt{GUARD}):} The \textit{During}, \textit{Starts}, and \textit{Finishes} relations can be defined by nesting $\mathcal{I}_A \subset \mathcal{I}_B$, which corresponds to the ${\texttt{GUARD}}$ operator. Fine-grained temporal relations under \texttt{GUARD} like \textit{Starts} and \textit{Finishes} can be represented by adding constraints to the corresponding validity gates.

\paragraph{(3) The Identity Group (\texttt{SYNC}):} The \textit{Equal} relation is defined by temporal equivalence $\mathcal{I}_A = \mathcal{I}_B$, corresponding to the ${\texttt{SYNC}}$ operator.

\paragraph{(4) The Inverse Group:} The \textit{After}, \textit{Contains}, and \textit{Overlapped-by} relations can be expressed by swapping the arguments of ${\texttt{SEQ}}$ and ${\texttt{GUARD}}$ (e.g., ${\texttt{SEQ}}(n_B, n_A)$ resolves \textit{After}).


\paragraph{\textit{Conclusion:}} Since all 13 relations map to parameterizations of $\mathcal{O}_{\text{core}}$ or their constructive inverses, the set is operationally complete. \hfill \QEDB

\section{Implementation Details}
\label{app:imp_details}

We summarize the prompts used by \textsc{SELA} below. Runtime-dependent inputs are represented by placeholders for readability.

\begin{promptbox}{ELT Schema Parser}
You are a Signal Logic Architect for time-series event detection using
Event Logic Trees (ELTs).

# Input
event_name: <event_name>
channels: <channels>
event_specification: <description>

# Task
Compile the natural-language event specification into a binary ELT.

Use the structure:
    Root = SEQ(Pre, SEQ(Main, Post))

where Main is the core event phase. If Pre or Post is unspecified,
use an observable background/holding state rather than inventing an event.

Primitive nodes:
- use exactly one available signal channel;
- describe an observable temporal morphology over an interval;
- have unique aliases.

Composite nodes use one of:
- SEQ(A,B): A is followed by B;
- SYNC(A,B): A and B substantially overlap;
- GUARD(A,B): A occurs within B;
- OR(A,B): either morphology may occur.

# Constraints
- Use a strict binary tree with unique aliases.
- Keep the tree compact while preserving event semantics.
- Prefer SYNC for simultaneous cross-channel patterns.
- Do not use purely negative or unobservable primitives.
- Do not duplicate equivalent morphology.

Validate the schema using `submit_schema`, refine it if needed, and
return the final schema with root_alias="Root" using `submit_result`.
\end{promptbox}

\begin{promptbox}{SELA Event Inspector}
You are an expert multivariate time-series analyst using an
Event Logic Tree (ELT).

# Input
candidate ELT(s): <tree_information>
dataset description: <description>
time series: accessed through visualization tools

# Task
Ground each candidate ELT in the observed time series by assigning
intervals and confidence scores to its primitive nodes.

# Tools
- `view_full`: inspect the full multivariate signal.
- `view_window`: inspect a candidate region in detail.
- `instantiate`: assign primitive intervals and confidence scores.
- `compare_trees`: compare candidate ELTs when multiple classes are loaded.

# Procedure
1. Inspect the full signal and identify plausible event regions.
2. Zoom into candidate regions before fixing boundaries.
3. Instantiate the relevant Pre, Main, and Post primitives.
4. Verify that each assigned interval matches its primitive morphology.
5. Refine intervals or confidence scores when ELT constraints are violated.
6. For multiple candidate classes, compare the normalized root confidence
   (`mu_norm`) and select the best-supported ELT.

# Grounding Rules
- SEQ: child phases must form a coherent temporal sequence.
- SYNC: child intervals should substantially overlap.
- GUARD: the guarded child must lie within its parent interval.
- OR: only supported branches need to be instantiated.

A primitive description must hold over its entire assigned interval.
Tighten boundaries when only a sub-region matches the morphology.

Use conservative and consistent confidence calibration:
0.9--1.0 strong match;
0.7--0.9 good match;
0.4--0.7 partial match;
0.1--0.4 weak match;
0 unsupported.

Do not inflate confidence to favor a particular event class.
Return a concise summary of the decisive evidence using `submit_comment`.
\end{promptbox}

\begin{promptbox}{Parallel Single-Tree Inspector}
When parallel inspection is enabled, each inspector receives only one
candidate event class and its ELT.

Instantiate the tree following the same grounding, operator, boundary,
and confidence rules above. Other event classes are evaluated independently.

A low normalized root confidence is the correct output when the observed
signals do not support the assigned event class. The final prediction is
obtained by comparing the independently computed `mu_norm` scores.
\end{promptbox}

\begin{promptbox}{Shared ReAct Instruction}
Before each tool call, briefly state what is currently observed,
what will be inspected or modified next, and why the action is useful.
\end{promptbox}

\section{Climate Hazard Construction Details}
\label{app:noaa}

\cam{Historical weather-event records are collected from heterogeneous sources and may not always provide fully reliable event annotations. NOAA notes that some external reports in the NWS Storm Events database may remain unverified because of time and resource constraints, and therefore ``does not guarantee the accuracy or validity of the information''\footnote{\url{https://www.ncei.noaa.gov/stormevents/faq.jsp}}. Detecting historical events directly from time series data can therefore provide additional evidence for checking and improving existing event records. To construct the Climate Hazard benchmark for this setting, we first select hazard types suitable for multivariate interval detection, then align the corresponding NWS records with ISD station observations and filter low-quality samples, and finally ask data scientists to inspect the signals and calibrate the event intervals.}

\subsection{Event Type Selection Rationale}
\label{app:event_type}

\cam{The Storm Events database contains dozens of hazard types, but not all of them are suitable for our task. We select \textit{Dense Fog}, \textit{Extreme Cold}, and \textit{Frost Freeze} based on the following considerations.}

\paragraph{Event frequency.}  Rare hazard types yield too few episodes to construct a dataset of meaningful scale, and calibrating near-singular events offers limited practical value.  \cam{\textit{Dense Fog}, \textit{Extreme Cold} and \textit{Frost Freeze}} are among the more frequently recorded hazards in the continental US, striking a balance between data availability and operational relevance.

\paragraph{ISD data quality and coverage.}  NWS event zones and ISD station networks are independent, so a recorded event does not guarantee a co-located station with adequate coverage. Many hazard types are further hindered by their spatially concentrated impact area: when a station lies outside the event core, the recorded signal may be too weak to be representative.

\paragraph{Event duration.}  Short-lived hazards such as \textit{Thunderstorm Wind} or \textit{Lightning} last only minutes which at or below the resolution of both ISD records and NWS timestamps, reducing ground-truth boundaries to near-point annotations unsuitable for interval detection. \cam{All} selected types persist over hours to days.

\paragraph{Multivariate necessity.}  Some hazards are detectable by a single threshold on one channel (e.g., \textit{Excessive Heat} reduces to a sustained temperature exceedance), making multivariate knowledge-guided detection unnecessary.

\subsection{Candidate Sample Construction}
\label{app:noaa-pipeline}

\paragraph{Event–station alignment.}
Each NWS event is associated with a geographic zone rather than a point station. For every event, we identify candidate ISD stations within or near the corresponding event zone whose active date ranges cover the event period. We retain only FM-15 (METAR) reports to ensure consistent observation timing and measurement fields.

\paragraph{Episode deduplication.}
A single meteorological episode may generate multiple NWS records across affected counties or zones. We group records belonging to the same episode and retain the station with the highest observation quality. Stations are ranked by
\begin{equation}
q = \frac{\bar{n}_c}{1 + \Delta},
\end{equation}
where $\bar{n}_c$ is the mean observation count per channel and $\Delta$ is the maximum inter-observation gap in minutes.

\paragraph{Quality filtering.}
Four filters are applied before a candidate is accepted: (1) $\geq$5 non-NaN readings per channel; (2) no gap exceeding 120 consecutive minutes; (3) $\geq$3 observations in each of the pre- and post-event sub-windows; (4) at least one observation with \texttt{VIS\_km} $< 5.0$ for \textit{Dense Fog}; \texttt{TEMP\_C} $< {-10}$\,°C / \texttt{WIND\_CHILL} $< {-20}$\,°C for \textit{Extreme Cold}; \cam{\texttt{TEMP\_C} $\leq 2.0$,°C for \textit{Frost Freeze}}

\paragraph{Event ratio constraint.}
Let $r = (t_{\mathrm{end}} - t_{\mathrm{start}}) / L$ denote the fraction of the window occupied by the GT event, where $L$ is the window length. Accepted ranges are $r \in [0.10, 0.80]$ for \textit{Dense Fog} as well as \cam{\textit{Frost Freeze}}, and $r \in [0.20, 0.80]$ for \textit{Extreme Cold}. Where ISD station records do not extend to the full target window on either side, the available data is retained as-is; no zero-padding or forward-filling beyond the last observation is applied.

\subsection{Event Description and Boundary Calibration}

\paragraph{Event descriptions.}
\cam{We construct textual descriptions of the selected hazard types from official NWS documentation and the American Meteorological Society Glossary%
\footnote{\url{https://www.weather.gov/media/directives/010_pdfs_archived/pd01016005c.pdf}}%
\footnote{\url{https://glossary.ametsoc.org/}}%
, with additional guidance for \textit{Extreme Cold} from the NWS Wind Chill Temperature Index%
\footnote{\url{https://www.weather.gov/media/ajk/brochures/Wind_Chill_Temperature_Index.pdf}}.
The descriptions summarize the relevant signal morphology, relationships among meteorological variables, and their temporal evolution. These descriptions provide domain knowledge for both language-guided detection and subsequent expert inspection.}

\paragraph{Manual signal inspection and calibration.}
\cam{The original event intervals are initialized from the NWS Storm Events database. As discussed above, these records are not always fully reliable and may therefore require further verification. Moreover, even when an event record is accurate at the geographic-zone level, its reported timestamps may not precisely match how the event appears at an individual ISD station. We therefore ask two annotators with climate-science backgrounds to independently inspect each candidate time series together with its event description. Samples with abnormal signals are removed, and the boundaries of the remaining events are manually adjusted when necessary to better match the observed meteorological evidence.}

\paragraph{Cross-verification.}
\cam{The two annotators independently verify the event category and boundaries. At the event-category level, their agreement is 100\%. For boundary annotations, we use an event-specific disagreement threshold $T$ that reflects the characteristic duration and formation/dissipation behavior of each hazard. When the two annotations differ by no more than $T$, their mean is used as the final boundary; otherwise, the annotators review the sample jointly and resolve the disagreement through discussion. We use $T=1$ hour for \textit{Dense Fog}, $T=4$ hours for \textit{Extreme Cold}, and $T=2$ hours for \textit{Frost Freeze}.
Across the climate hazard data, the pre-discussion boundary agreement rate is 91\%.}

\section{ELT Parse Quality Evaluation}
\label{app:elt-parse-quality}

\subsection{Evaluation Dimension Details}
Each valid parse is scored against the ground-truth ELT and the original event description by two independent VLM judges, Claude Opus-4.6 and GPT-5.5 (thinking mode), drawn from different model families to reduce scoring bias. \cam{Each judge evaluates the same parse three times.} \cam{To mitigate position bias, we randomise whether the LLM parse or the ground-truth tree is shown first in each evaluation.} Scores on a 0--5 scale across four dimensions are averaged and normalised to $[0, 1]$. The two judges show strong agreement (Spearman $\rho = 0.86$, $p < 0.001$ across all scored parses), and VLM scores on a held-out subset of 30 parses correlate strongly with independent human annotations ($\rho = 0.83$), \cam{supporting the reliability of the evaluation.}

\subsection{Judge Prompt}

\begin{promptbox}{LLM Judge Instruction}
# Role & Objective
You are an expert judge evaluating the quality of Event Logic Trees (ELTs) 
automatically parsed by LLMs from natural-language event descriptions. You 
will be given:
1. A ground-truth ELT (GT) parsed by a human expert
2. An LLM-parsed ELT for the same event
3. A side-by-side visualisation of the two trees
4. The original natural-language event description

Your task is to score the LLM-parsed ELT on four dimensions (0-5 scale).

# Scoring Dimensions

## Structural Fidelity (0-5)

How similar is the LLM tree structure to the GT in terms of operator 
types and tree topology?
- 5: Identical operator types and tree shape, or logically equivalent 
     (e.g., different grouping but same logical meaning)
- 4: Minor differences in nesting or operator choice, but overall 
     structure preserved
- 3: Similar but with extra nesting, wrong operators, or different 
     grouping that changes logical meaning
- 1-2: Major structural differences
- 0: Completely different structure

## Primitive Coverage (0-5)

Are all ground-truth primitives accounted for in the LLM parse? A 
primitive may be covered at a different granularity (e.g., split into 
sub-stages) as long as it is logically equivalent.
- 5: All GT primitives fully covered (exactly or logically equivalent)
- 3: Most GT primitives present; minor omissions
- 0: Major GT primitives missing

## Primitive Accuracy (0-5)

Are all LLM-parsed primitives grounded in the original event description? 
This dimension penalises hallucinated or fabricated content, regardless 
of how many GT primitives were covered.
- 5: Every primitive has clear grounding in the event description; 
     no hallucinated channels or fabricated signal names
- 3: Minor extras or slightly imprecise channel references
- 0: Many hallucinated or fabricated primitives

## Semantic Alignment (0-5)

How well does each primitive's description match the corresponding part 
of the original event description?
- 5: Precise and accurate descriptions for all primitives
- 3: Roughly correct but imprecise or incomplete
- 0: Wrong or contradicts the event definition

# Important Notes
- Logical equivalence counts as a match for both Structural Fidelity 
  and Primitive Coverage (e.g., different granularity, reordered 
  children with same semantics)
- Primitive Accuracy and Coverage are orthogonal: a parse with one 
  perfectly grounded primitive but all others missing scores 5 on 
  Accuracy but 0-1 on Coverage

# Output Format
Return a JSON object with the following structure:
{
  "structural": <float 0-5>,
  "coverage": <float 0-5>,
  "accuracy": <float 0-5>,
  "semantic": <float 0-5>,
  "rationale": "<brief explanation of scores>"
}

\end{promptbox}

\section{Evaluation Metrics}
\label{app:metrics}

\paragraph{Why F1-IoU is insufficient for Climate Hazard.}
NWS event times reflect zone-level conditions rather than point-station measurements, and physical event boundaries are inherently gradual (e.g., fog dissipates over 30--60\,min). We observe F1@IoU-0.5 is much larger than F1@IoU-0.9 across all models, confirming that hard IoU thresholds collapse meaningful performance differences: a model that detects the correct event with a 20\% boundary error receives the same zero score as one that misses the event entirely.

\paragraph{Coverage-based metrics.}
We adopt a coverage-based formulation following~\citet{NEURIPS2018_8f468c87},
instantiated with flat positional bias and no cardinality penalty.
For a matched pair (GT interval $[s_g, e_g]$, predicted interval
$[s_p, e_p]$), let
\begin{equation}
  o = \max\bigl(0,\min(e_g,e_p)-\max(s_g,s_p)\bigr).
\end{equation}
and define coverage recall and precision as
\begin{equation}
  C_g = \frac{o}{e_g - s_g}, \qquad
  C_p = \frac{o}{e_p - s_p}.
\end{equation}
These correspond to range-based recall and precision respectively
under the framework of~\citet{NEURIPS2018_8f468c87}.
Their harmonic mean gives the neutral aggregate:
\begin{equation}
  \text{F1}^{\texttt{Cov}} = \frac{2\,C_g\,C_p}{C_g + C_p}.
\end{equation}
Intervals are matched one-to-one (greedy, by descending
F1$^{\texttt{Cov}}$) within \cam{each event class};
unmatched GT intervals contribute $C_g = 0$.
\cam{The resulting scores are computed independently for each event
class and macro-averaged across classes. We denote the reported
macro-averaged metric as $\mathcal{M}$F1$^{\texttt{Cov}}$.}

\paragraph{Precision-weighted metric F0.5$^{\texttt{Cov}}$.}
F1$^{\texttt{Cov}}$ treats $C_g$ and $C_p$ equally.
We argue that precision deserves higher weight in this setting for
two complementary reasons.
First, NWS boundaries represent gradual physical transitions; $C_g$
therefore penalises boundary errors that fall within the inherent
uncertainty of the annotation itself, making it a noisier signal
than $C_p$.
Second, in operational hazard detection a false alarm---predicting
an event outside its true temporal extent---triggers unnecessary
emergency response, whereas a conservative boundary miss is less
disruptive.
We therefore additionally report F0.5$^{\texttt{Cov}}$
($\beta=0.5$), which weights precision twice as heavily as
recall.
Following~\citet{NEURIPS2018_8f468c87}, this corresponds to their
\textsc{Reward-Low-FP} application profile:
\begin{equation}
  \text{F0.5}{}^{\texttt{Cov}} =
    \frac{1.25\,C_g\,C_p}{0.25\,C_p + C_g}.
\end{equation}
F1$^{\texttt{Cov}}$ is retained as the unweighted baseline.
\cam{Analogously, F0.5$^{\texttt{Cov}}$ is computed per event class
and macro-averaged across classes, yielding the reported
$\mathcal{M}$F0.5$^{\texttt{Cov}}$ metric.}

\section{Numeric and VL-Time Baseline Prompts}
\label{app:baseline-prompts}

\subsection{Numeric Baseline}

\begin{promptbox}{Numeric Direct TS Event Detection}
You are an expert in time-series analysis.

# Input
event_types: <event_types>
dataset_description: <description>
time_series: <raw time-series values in CSV format>

# Task
Identify the temporal subsequence corresponding to each target event
type and predict its start and end indices.

For every event type, output exactly one best candidate interval.
If an event appears absent, still provide the most plausible candidate
with a low confidence score. The event type believed to be present
should receive the highest confidence.

# Output
Return ONLY a JSON array:

[
  {
    "className": "<event_name>",
    "start": <integer>,
    "end": <integer>,
    "confidence": <float in [0,1]>
  }
]

Output one entry per event type, with no additional text.
\end{promptbox}

\subsection{VL-Time Baseline}

\begin{promptbox}{VL-Time Visual TS Event Detection}
You are an expert in signal processing and time-series analysis.

# Input
event_types: <event_types>
dataset_description: <description>
time_series_info: <plot metadata>
time_series_visualization: <full-range multichannel plot>

# Task
Inspect the visualization and identify the temporal subsequence
corresponding to each target event type, with precise start and end
indices.

For every event type, output exactly one best candidate interval.
If an event appears absent, still provide the most plausible candidate
with a low confidence score. The event type believed to be present
should receive the highest confidence.

# Output
Return ONLY a JSON array:

[
  {
    "className": "<event_name>",
    "start": <integer>,
    "end": <integer>,
    "confidence": <float in [0,1]>
  }
]

Output one entry per event type, with no additional text.
\end{promptbox}

\subsection{Few-Shot Demonstrations}

\begin{promptbox}{Numeric Few-Shot Demonstration}
# Annotated Example

The following time series comes from the same dataset and is accompanied
by its ground-truth events. Use the example to learn the output format
and the numeric signatures of the event types.

# Time Series Data
<raw time-series values in CSV format>

# Ground-Truth Output
[
  {
    "className": "<event_name>",
    "start": <integer>,
    "end": <integer>,
    "confidence": 1.0
  }
]
\end{promptbox}

\begin{promptbox}{VL-Time Few-Shot Demonstration}
# Annotated Example

The following visualization comes from the same dataset and is
accompanied by its ground-truth events. Use the example to learn the
output format and the visual morphology of the event types.

# Time Series Info
<plot metadata>

# Time-Series Visualization
<full-range multichannel plot>

# Ground-Truth Output
[
  {
    "className": "<event_name>",
    "start": <integer>,
    "end": <integer>,
    "confidence": 1.0
  }
]
\end{promptbox}

\begin{table}[t]
\centering
\scalebox{0.82}{
\begin{tabular}{@{}llrrrr@{}}
\toprule
& & \multicolumn{2}{c}{Zero-shot} & \multicolumn{2}{c}{Few-shot} \\
\cmidrule(lr){3-4}\cmidrule(l){5-6}
Set & Method & Tokens & Cost & Tokens & Cost \\
\midrule

\multirow{6}{*}{\textbf{PT}}
 & \textsc{SELA} (GPT-5)      & 233.6 & 0.623 & -   & -   \\
 & \textsc{SELA} (GPT-4.1)    & 125.9 & 0.189 & -   & -   \\
 & Num. (GPT-5)            &  33.6 & 0.081 &  91.9 & 0.143 \\
 & Num. (GPT-4.1)          &  29.2 & 0.059 &  92.2 & 0.185 \\
 & VL-T. (GPT-5)            &   5.1 & 0.033 &   6.7 & 0.031 \\
 & VL-T. (GPT-4.1)          &   2.3 & 0.005 &   4.7 & 0.010 \\
\midrule

\multirow{6}{*}{\textbf{CH}}
 & \textsc{SELA} (GPT-5)      & 295.4 & 0.638 & -   & -   \\
 & \textsc{SELA} (GPT-4.1)    & 188.5 & 0.296 & -   & -   \\
 & Num. (GPT-5)            &  39.6 & 0.091 & 142.3 & 0.203 \\
 & Num. (GPT-4.1)          &  35.0 & 0.071 & 139.5 & 0.280 \\
 & VL-T. (GPT-5)            &   5.5 & 0.033 &   8.2 & 0.033 \\
 & VL-T. (GPT-4.1)          &   3.0 & 0.008 &   6.5 & 0.014 \\
\bottomrule
\end{tabular}
}
\caption{Average per-sample token usage (K) and inference cost (\$) for LLM-based methods. The slightly lower few-shot cost of VL-Time with GPT-5 on Pressure Test results from reduced output-token usage despite the longer input prompt.}
\label{tab:token_cost}
\end{table}

\section{Token Budget}
\label{app:token_budget}

\cam{Table~\ref{tab:token_cost} reports the average per-sample token usage and inference cost of the LLM-based methods. \textsc{SELA} uses more tokens than Numeric and VL-Time because it performs iterative ELT reasoning, signal inspection, and confidence estimation through multiple agent interactions. Despite this additional computation, its average inference cost remains below \$0.65 per sample in all settings. Notably, \textsc{SELA} with GPT-4.1 uses a token budget comparable to few-shot Numeric, while achieving substantially better detection and localization performance. Numeric requires fewer tokens in the zero-shot setting, while VL-Time is the most token-efficient baseline because it reasons directly over compact time-series visualizations. Few-shot prompting increases the token usage of both baselines by including annotated demonstrations.}

\end{document}